\documentclass[11pt]{article}

\usepackage[final]{acl}

\usepackage{times}
\usepackage{latexsym}
\usepackage[T1]{fontenc}
\usepackage[utf8]{inputenc}
\usepackage{microtype}
\usepackage{inconsolata}

\usepackage{lineno}

\usepackage{amsmath}
\usepackage{amssymb}
\usepackage{pifont}
\usepackage{multirow}
\usepackage{multicol}
\usepackage{booktabs}
\usepackage{graphicx}
\usepackage{threeparttable}
\usepackage{enumitem}

\usepackage{subcaption}

\usepackage{makecell}

\usepackage[most]{tcolorbox}

\usepackage{booktabs}
\usepackage{tabularx}
\usepackage{array}
\usepackage{makecell}
\usepackage{ragged2e}
\usepackage{xcolor}
\usepackage{listings}

\newcolumntype{Y}{>{\RaggedRight\arraybackslash}X}

\lstdefinestyle{AppendixJson}{
    basicstyle=\ttfamily\footnotesize,
    breaklines=true,
    breakatwhitespace=true,
    columns=fullflexible,
    keepspaces=true,
    frame=single,
    showstringspaces=false
}

\begin{document}
% \nocite{*}
\title{CBRS: Cognitive Blood Request System with Bilingual Dataset and Dual-Layer Filtering for Multi-Platform Social Streams}

\author{
\textbf{Anik Saha}$^1$$^*$ \hspace{1em}
\textbf{Mst. Fahmida Sultana Naznin}$^1$$^*$ \hspace{1em}
\textbf{Zia Ul Hassan Abdullah}$^1$ \\
\textbf{Anisa Binte Asad}$^1$ \hspace{1em}
\textbf{K.\ G.\ Subarno Bithi}$^1$ \hspace{1em}
\textbf{A.\ B.\ M.\ Alim Al Islam}$^1$ \\
$^1$Bangladesh University of Engineering and Technology, Dhaka, Bangladesh \\
\texttt{aaniksahaa.2001@gmail.com, nazninfahmidasultana@gmail.com, 2005037@ugrad.cse.buet.ac.bd} \\
\texttt{anisabinteasad134@gmail.com, 2011013@mme.buet.ac.bd, alim\_razi@cse.buet.ac.bd} \\
$^*$These authors contributed equally and are listed alphabetically.
}

% \author{Mst. Fahmida Sultana Naznin$^1$ \quad Adnan Ibney Faruq$^1$ \quad Mostafa Rifat Tazwar$^1$ \quad Md Jobayer \quad Md. Mehedi Hasan Shawon \quad Md Rakibul Hasan$^5$ \\
%         $^1$Bengalidesh, Dhaka , Bengalidesh \\
%         $^5$Curtin University, Bentley WA 6102, Australia \\
%         \texttt{nazninfahmidasultana@gmail.com, adnanibneyf@gmail.com, 2005020@ugrad.cse.buet.ac.bd
%         \\rakibul.hasan@curtin.edu.au}}

\maketitle

\begin{abstract}

Urgent blood donation seeking posts and mesages on social media often go unnoticed due to the
overwhelming volume of daily communications. Traditional app-based systems,
reliant on manual input, struggle to reach users in low-resource settings, delaying critical responses. To address this, we introduce the Cognitive Blood
Request System (CBRS), a multi-platform framework that efficiently filters and
parses blood donation requests from social media streams using a cost-efficient
dual-layered architecture. To do so, we curate a novel dataset of 11K parsed blood donation request messages in Bengali, English, and transliterated Bengali, capturing the linguistic diversity of real social media communications. The inclusion of adversarial negatives further enhances the robustness of our model. CBRS achieves an impressive 99\% accuracy and precision in filtering, surpassing benchmark methods. In the parsing task, our LoRA finetuned Llama-3.2-3B model achieves 92\% zero-shot accuracy surpassing the base model by 41.54\% and exceeding the few-shot performance of GPT-4o-mini, gemini-2.0-flash and other LLMs while resulting in a 35$\times$ reduction in input token usage. This work lays a robust foundation for scalable, inclusive information extraction in time-sensitive, object-focused tasks. Our code, dataset, and trained models are publicly available at \url{https://github.com/aaniksahaa/CBRS}.

\end{abstract}

\section{Introduction}

In the digital era, social networking sites (SNSs) have fueled the rapid growth
of online communities, with millions of posts shared daily \cite{social-media-use-2021}. Amid emergencies, users increasingly rely on these platforms to broadcast urgent blood
donation needs, seeking to connect with potential donors \cite{saudi-study}. However, without efficient automated systems, such posts often remain buried within users’ immediate
social circles, limiting their reach \cite{ebdr-twitter}. The unstructured and scattered nature
of social media communications poses significant challenges for extracting critical information and efficiently disseminating these requests \cite{ABBASI2018892, IE-social-2022}.

\begin{figure}[t]
    \centering
    \includegraphics[width=1.0\linewidth]{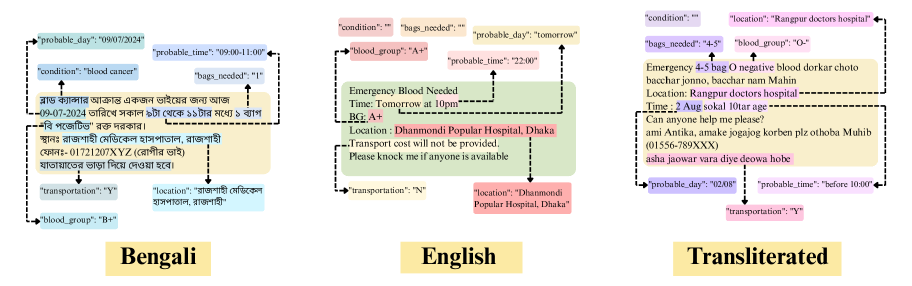} % replace with your figure
    \caption{Bilingual parsing methodology from Bengali-English-Transliterated Bengali blood request corpora}
    \label{fig:message}
\end{figure}

\begin{figure}[t]
    \centering
    \begin{subfigure}{0.15\textwidth}
        \includegraphics[width=\textwidth, page=1]{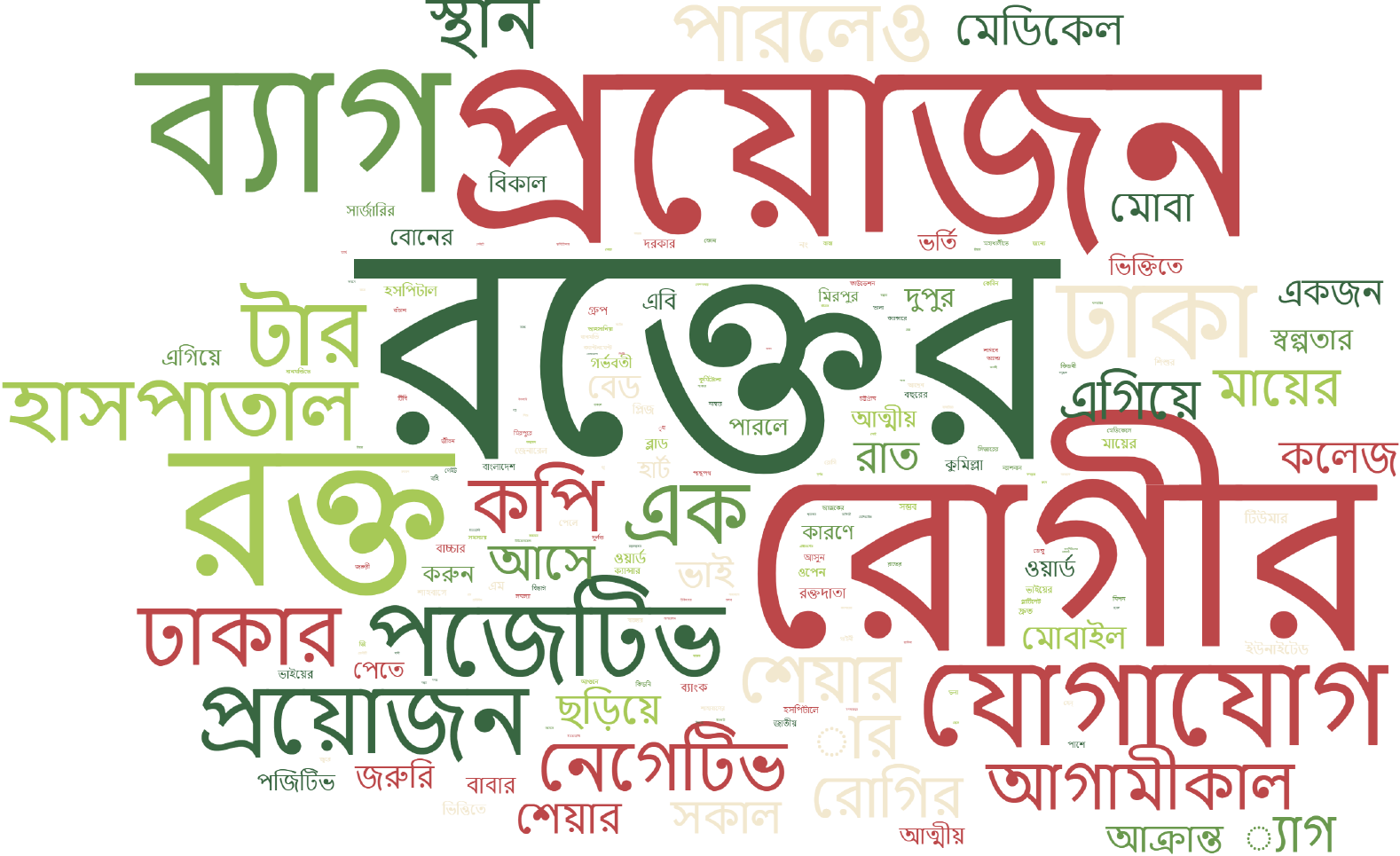}
        \caption{Bengali}
    \end{subfigure}
    \hfill
    \begin{subfigure}{0.15\textwidth}
        \includegraphics[width=\textwidth, page=1]{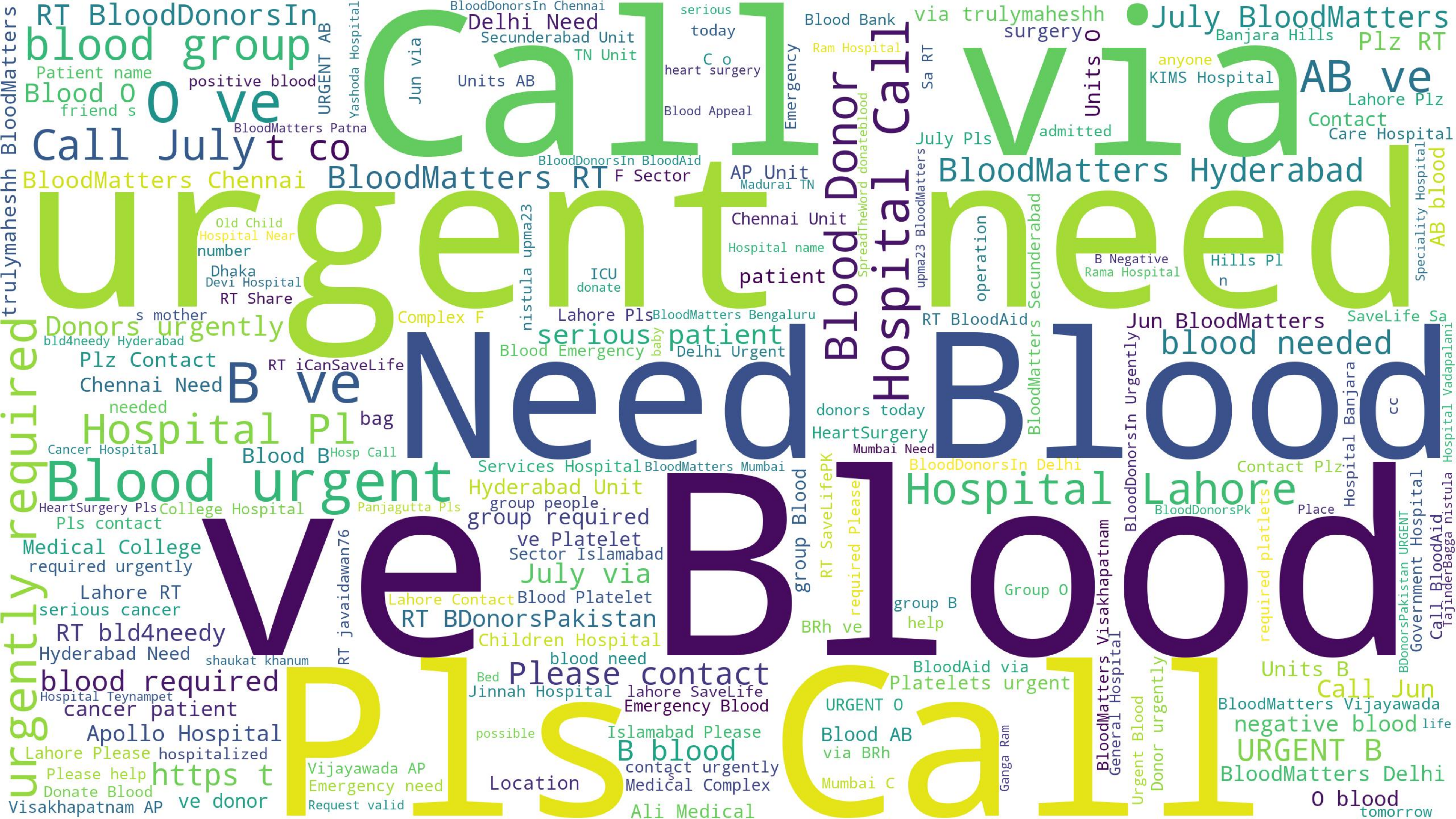}
        \caption{English}
    \end{subfigure}
    \hfill
    \begin{subfigure}{0.15\textwidth}
        \includegraphics[width=\textwidth, page=1]{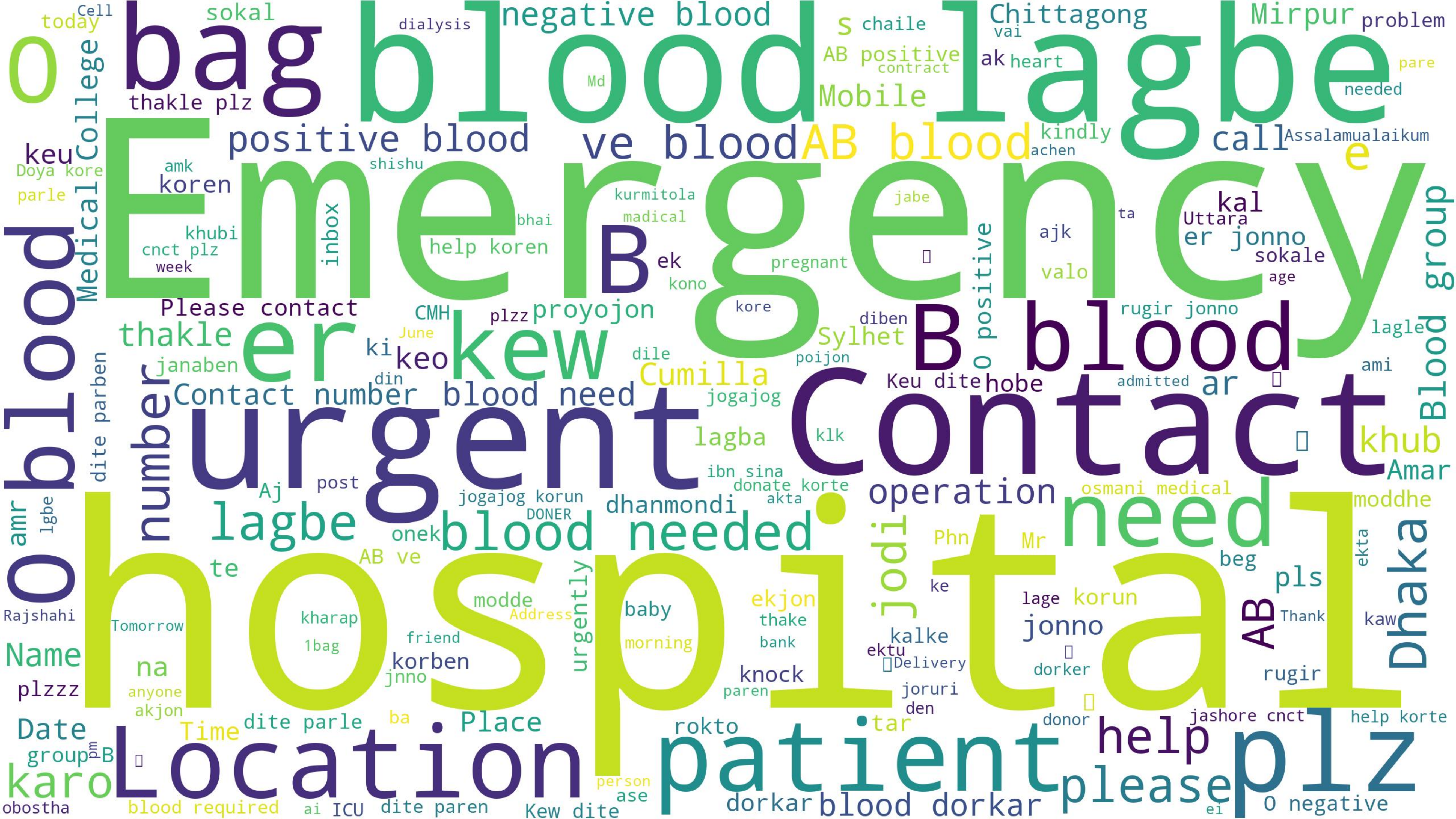}
        \caption{Transliterated Bengali}
    \end{subfigure}
   
    \caption{Wordcloud of top keywords in CBRS dataset}
    \label{fig:wordcloud}  % Move label here
\end{figure}

A key limitation in filtering and parsing such messages in a multilingual setting lies in the limited availability of datasets for low-resource languages such as Bengali. Most state-of-the-art Natural Language Processing architectures rely on large-scale annotated corpora, which are scarcely available for low-resource languages. These languages often feature complex morphosyntactic structures, diverse dialectal variations, and unique linguistic phenomena as shown in Figure \ref{fig:message} that are underrepresented in existing multilingual pre-trained models, limiting effective generalization and transfer learning. Although there are data sets for the classification of disaster and emergency requests \cite{ebdr-twitter, crisis-bench}, none specifically include Bengali or transliterated Bengali. To our knowledge, we introduce the first bilingual data set comprising requests for blood donation in English, Bengali, and transliterated Bengali. Figure \ref{fig:wordcloud} shows the wordcloud of our dataset.

Developing a reliable solution for accurate detection and effective dissemination of emergency blood donation requests to potential donors poses several critical challenges. Firstly, the volume of incoming messages and social media posts is often overwhelming, but only a small fraction of these messages represent actual blood donation requests. Furthermore, in case of classifying such requests, false negatives are much more detrimental than false positives, since the former implies ignoring an urgent request while the latter only adds a little more load to subsequent processing. Although there is existing work on this disaster and emergency related message classification \cite{le2022disastertweetsclassificationusing, POWERS2023100164, 10.1007/978-3-030-92537-6_15}, they often overlook this asymmetric nature of the problem. Secondly, merely detecting whether a message is asking for blood donation is insufficient to determine which donors to notify to maximize the likelihood of a rapid response. An automated parsing of such free-form texts is essential to extract the key information in a structured format. However, previous studies have focused mainly on detection \cite{cheng-etal-2025-xformparser, wan2024omniparserunifiedframeworktext}, leaving a gap in designing efficient and scalable parsing solutions. Thirdly, for such a system to be viable in real-world deployment, it must balance speed and accuracy, which present conflicting design constraints. For instance, using a naively trained lightweight Machine Learning (ML) model for the classification leaves a chance of a higher number of false negatives, while using a large language model (LLM) entirely for this classification task will not be scalable due to high inference times and costs given the volume of incoming data.

To address these challenges, we propose a cost-efficient dual-layered filtering architecture to identify blood donation requests from large message pools effectively, coupled with a cost-efficient LLM for rapid and accurate parsing of free-form text requests into a predefined structured format. Our key contributions are as follows:
\begin{itemize}
    \item We present a novel parsed bilingual dataset consisting of 11K Bengali-English-Transliterated Bengali blood donation requests sourced from social media. This dataset is further enriched with curated adversarial negatives and fragments from publicly available datasets.
    \item We present the Cognitive Blood Request System (CBRS), which integrates a cost-efficient dual-layered filtering architecture designed to efficiently detect blood donation requests taking into account the asymmetric class weighting. 
    \item We train a LoRA finetuned Llama-3.2-3B model for parsing and compare its performance with other open and closed-weight LLMs in zero-shot and few-shot settings. 
    \item We benchmark CBRS against existing filtering and parsing methods in terms of both performance and computational complexity. In a separate human evaluation study across 30 active Telegram and Discord groups with diverse demographics, we assess the real-world effectiveness of our approach and identify the key factors influencing user satisfaction. 
\end{itemize}
\begin{figure*}[t]
    \centering
    \includegraphics[width=0.73\textwidth]{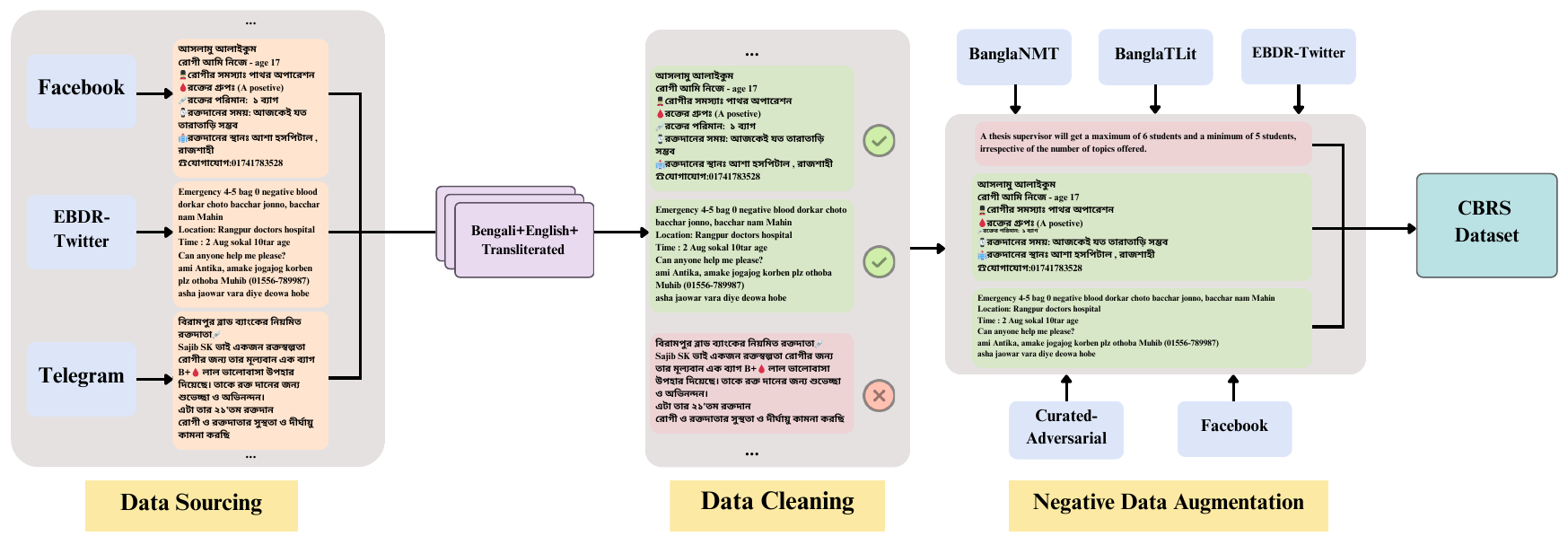}
    \caption{Data sourcing process of CBRS: positive samples are collected from Facebook, EBDR-Twitter, and Telegram, followed by cleaning and augmentation with negative samples from BanglaNMT, BanglaTLit, EBDR-Twitter, Facebook, and curated adversarial samples.}
    \label{fig:database}
\end{figure*}

\section{Related Work}\label{sec_rl}
\paragraph{Information Extraction from Social Media}Social media is vital for real-time updates during emergencies, but its unstructured and noisy nature makes extracting actionable insights difficult. Recent advances in AI and NLP, especially LLMs, offer promising solutions. Belcastro et al. \cite{Belcastro2025multi} used LLMs to classify disaster-related content by emotion, sentiment, and topic, generating stakeholder-specific summaries \cite{Belcastro2025multi}. He and Hu developed an AI system combining NLP and geospatial visualization for effective monitoring \cite{he2025social}. Yin et al. proposed CrisisSense-LLM for multi-label classification of event type, informativeness, and aid relevance \cite{yin2024crisissense}. Shetty et al. achieved over 91\% accuracy using multimodal learning on social media text and images \cite{shetty2024disaster}. Hu et al. introduced a geo-knowledge-guided GPT for location extraction, outperforming traditional NER by 40\% \cite{hu2023geo}. Alharbi and Haq applied DistilBERT for tweet classification, with 92.42\% training and 82.11\% validation accuracy \cite{alharbi2024enhancing}. Mehmood et al. proposed a three-step method for classifying relevant posts, extracting locations, and topic modeling with high F1-scores \cite{mehmood2024named}. However, specific extraction of blood-related requests remains largely unexplored.
\paragraph{Low-Resource Language Dataset Curation}
Information extraction using LLMs is increasingly applied in disaster response. However, in low-resource languages like Bengali, curated and task-specific datasets are so few \cite{acl1hasan-etal-2020-low, fahim2024banglatlit} that they remain a major bottleneck. Mathur et al. \cite{ebdr-twitter} proposed a system to identify emergency blood donation requests on Twitter, highlighting the potential of social media mining for critical healthcare interventions. CrisisBench \cite{crisis-bench} aggregates past disaster datasets into a unified benchmark for informativeness and urgency prediction. CrisisMMD \cite{crisismmd2018}, an early multimodal dataset, integrates text and images from Twitter for disaster classification. Bengali.AI \cite{bengaliai2020} contributed handwritten and speech-text corpora. These resources aid multilingual crisis AI, though blood related Bengali datasets are still missing.

\section{Dataset}
To overcome the limitations of current Bengali transliteration datasets, our design centers on two key goals: developing a Bengali-English-Transliterated Bengali corpus for blood donation requests, and capturing the diverse texting styles in social media groups, including dialectal variations, slang, and abbreviations that helps create a rich understanding of how language evolves in online communication.

\subsection{Data Sourcing}
We source Bengali, English, and Transliterated Bengali messages from 15 public blood donation groups on Telegram and Facebook. In total, we present a dataset of 11K parsed emergency blood donation requests as shown in Table \ref{tab:sample_source_stats}.

\subsection{Data Cleaning}
After aggregating the data sources, we conduct extensive deduplication and also detect samples that are not directly associated with blood donation requests. Certain messages- such as expressions of willingness to donate (e.g., “I can donate A- blood in Dhaka. Please contact me if you're a recipient”) or post-donation acknowledgments -although structurally similar to positive instances, do not represent actual requests. We classify these as \textbf{hard negatives}: non-relevant samples that closely mirror the linguistic and contextual patterns of true positives. Since these samples may particularly introduce semantic ambiguity, we keep them in the negative portion of the dataset to improve the robustness of the classifier.

\begin{table}[h!]
\centering
\caption{Sample distribution across different sources}
\resizebox{\columnwidth}{!}{%
\begin{tabular}{@{}lllll@{}}
\toprule
% \textbf{Category} & \textbf{Source} & \textbf{Total Samples} & \textbf{Total Tokens} & \textbf{Average Tokens} \\ \midrule
\textbf{Category} & \textbf{Source} & \makecell[l]{\textbf{Total}\\\textbf{Samples}} & \makecell[l]{\textbf{Total}\\\textbf{Tokens}} & \makecell[l]{\textbf{Average}\\\textbf{Tokens}} \\ \midrule

\multirow{3}{*}{Positive} 
& Facebook       & 6321  & 1747772 & 276.50 \\
& EBDR-Twitter   & 3941  & 169290  & 42.96  \\
& Telegram       & 744   & 139948  & 188.10 \\ 
\cmidrule(lr){2-5}
& \textbf{Total} & \textbf{11006} & \textbf{2057010} & -- \\ \midrule

\multirow{5}{*}{Negative} 
& BengaliNMT           & 3194  & 236220  & 73.96  \\
& BengaliTLit          & 5000  & 773058  & 154.61 \\
& Curated-Adversarial & 600   & 26211   & 43.69  \\
& Facebook            & 250   & 92262   & 369.05 \\
& EBDR-Twitter        & 5851  & 222568  & 38.04  \\
\cmidrule(lr){2-5}
& \textbf{Total}      & \textbf{14895} & \textbf{1350319} & -- \\ 

\bottomrule
\end{tabular}%
}
\label{tab:sample_source_stats}
\end{table}

\subsection{Negative Data Augmentation}
The dataset includes both positive (1: blood donation needed) and negative (0: not related) samples, carefully labeled for classification. We leverage Bengali and English texts from the BengaliNMT \cite{acl1hasan-etal-2020-low}, Bengali and Transliterated Bengali texts from the BengaliTLit \cite{fahim2024banglatlit}. The hard negatives that were manually filtered out in the previous phase are included in the negative portion. We also include curated adversarial examples containing terms like "blood", "urgent", and "emergency" to enhance robustness. These adversarial samples are generated with Deepseek-V3 using the aforementioned hard-negative samples for few-shot prompting. We obtain a portion of negative samples from the EBDR dataset as well. An overview is provided in Table \ref{tab:sample_source_stats}. Table \ref{tab:pos_neg_language} summarizes the Bengali, English, and Transliterated sample distribution across both categories. Figure \ref{fig:database} shows the workflow of data curation.

\begin{table}[t]
\centering
\caption{Sample distribution across different languages}
\resizebox{\columnwidth}{!}{%
\begin{tabular}{@{}llllll@{}}
\toprule
% \textbf{Category} & \textbf{Language} & \textbf{Total Sample} & \textbf{Total Tokens} & \textbf{Total Average} \\ \midrule
\textbf{Category} & \textbf{Language} & \makecell[l]{\textbf{Total}\\\textbf{Samples}} & \makecell[l]{\textbf{Total}\\\textbf{Tokens}} & \makecell[l]{\textbf{Average}\\\textbf{Tokens}} \\ \midrule

\multirow{3}{*}{Positive} 
& Bengali  & 6163 & 1829929 & 296.92 \\ 
& English  & 4412 & 197030  & 44.66  \\ 
& Transliterated & 431  & 30051   & 69.72  \\ 
 \midrule

\multirow{3}{*}{Negative} 
& Bengali  & 4420 & 893583  & 202.17 \\ 
& English  & 7663 & 264333  & 34.49  \\ 
& Transliterated & 2812 & 192403  & 68.42  \\

\bottomrule
\end{tabular}%
}

\label{tab:pos_neg_language}
\end{table}

\section{Methodology}
Messages circulating on SNSs for blood donations are often unstructured which complicates the automation of donor matching based on complex criteria and impeding rapid responses. Due to social media clustering, these requests typically reach a limited audience, with potential donors frequently overlooking them amidst vast content. To address these challenges, our methodology incorporates three components: 1) a cost-optimized dual-layered filtering system for detecting blood-related requests in groups 2) structured message parsing with Few-shot prompting 3) efficient donor notifications using a multi-platform control system based on geo-location.

\subsection{Proposed Dual Layered Filtering (DLF)}
\paragraph{\textbf{Layer 1: TF-IDF-and-Asymmetrically Weighted LogReg Classifier}}
This model is designed to detect blood requests written in Bengali, English, and Transliterated Bengali, handling bilingual and mixed-language texts, which are critical for the linguistic diversity in our dataset. The architecture of our model follows a systematic transformation of raw text into a binary classification decision for filtration of blood requests from extensive streams. 
\paragraph{\textbf{Subword Tokenization and Embedding}}
An input message \( M = (w_1, w_2, ..., w_T) \), where \( w_t \) represents the \( t \)-th word in the message, each word \( w_t \) is decomposed into subword units \( S(w_t) = \{ s_1, s_2, ..., s_N \} \) to capture linguistic variations. Here, \( S(w_t) \) denotes the set of subwords corresponding to word \( w_t \). The subwords are mapped to dense embeddings using an embedding matrix \( E \), where \( E \in \mathbb{R}^{d \times |V|} \) is a trainable matrix of dimension \( d \) (embedding size) and vocabulary size \( |V| \):
$\mathbf{v}_{s_i} = E \cdot \mathbf{1}_{s_i}$
where \( \mathbf{1}_{s_i} \) is the one-hot encoding of subword \( s_i \). The final word embedding is obtained by averaging over its subwords:
\begin{equation}
    \mathbf{v}_{w_t} = \frac{1}{|S(w_t)|} \sum_{s \in S(w_t)} \mathbf{v}_s
\end{equation}
where \( \mathbf{v}_s \) is the embedding vector of subword \( s \).

\paragraph{\textbf{Message Representation and Feature Extraction}}
To derive a fixed-length message representation, we apply average pooling over all word embeddings:
\begin{equation}
    \mathbf{V} = \frac{1}{T} \sum_{t=1}^{T} \mathbf{v}_{w_t}
\end{equation}
where \( T \) denotes the total number of words in the message. This vector \( \mathbf{V} \) is then processed through a fully connected layer for feature extraction:
$\mathbf{z} = W \mathbf{V} + b$ where \( W \in \mathbb{R}^{m \times d} \) is a weight matrix, \( b \in \mathbb{R}^{m} \) is a bias term, and \( \mathbf{z} \) represents the transformed feature vector of dimension \( m \).

\paragraph{\textbf{Binary Classification}}
Finally, a softmax layer is employed to predict the binary class—whether the message is related to a blood donation request (\( y=1 \)) or not (\( y=0 \)). The probability distribution over classes is computed as:
\begin{equation}
    P(y = c | \mathbf{z}) = \frac{\exp(z_c)}{\sum_{j} \exp(z_j)}
\end{equation}
where \( z_c \) is the logit corresponding to class \( c \).

To address the high cost of false negatives in emergency blood donation request detection, we adopt a weighted binary cross-entropy loss that penalizes misclassified positive examples more heavily. This asymmetry ensures the model prioritizes recall in the first layer. The loss function is defined as:

$$
\mathcal{L} = - \alpha \, y \log P(y = 1 \mid \mathbf{z}) - (1 - y) \log P(y = 0 \mid \mathbf{z})
$$

where $y \in \{0, 1\}$ is the true label, and we empirically choose $\alpha = 12$.

This asymmetric weighting, however, increases the number of false positives in the first phase. But since the overall fraction of blood donation requests in a general message pool is usually low, it adds negligible overhead to the subsequent phase.

\begin{figure*}[t]
    \centering
    \includegraphics[width=0.7\textwidth]{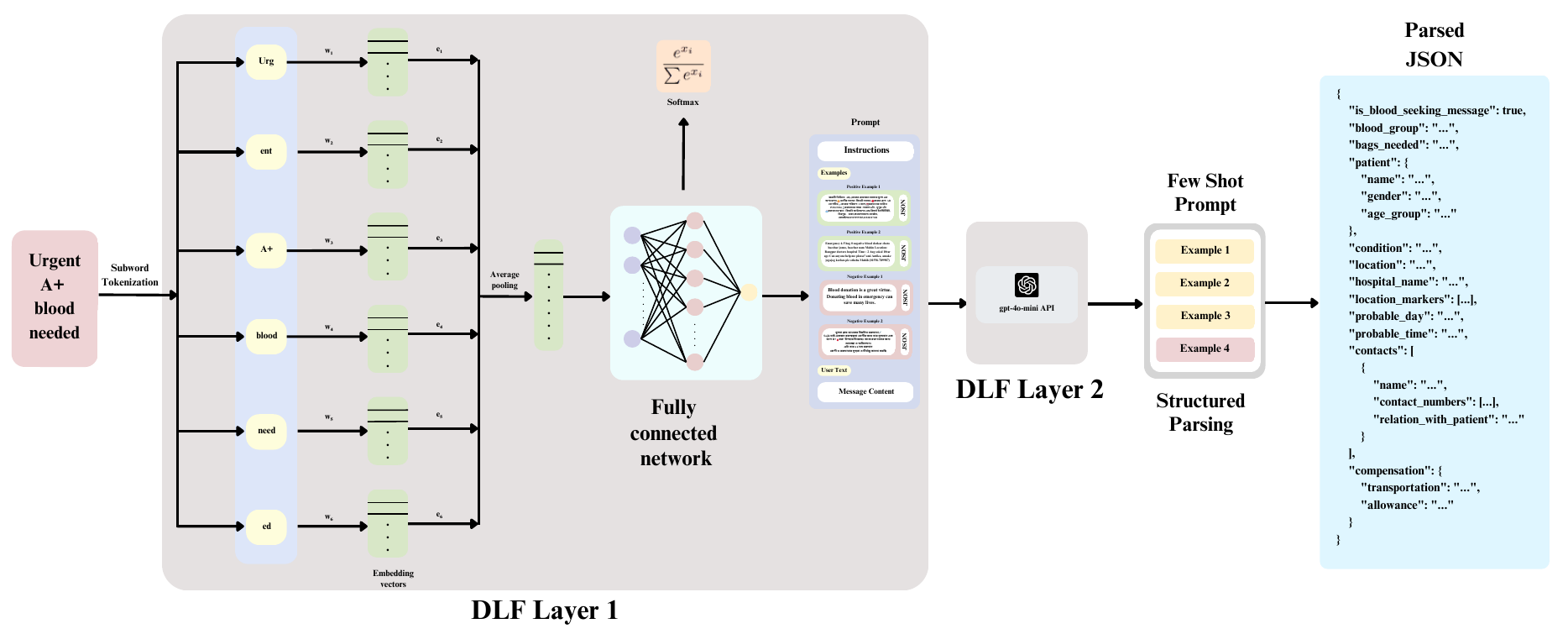}
    \caption{Dual-layered filtering and structured parsing architecture of CBRS, where raw messages undergo tokenization, pooling, and classification, followed by LLM–based filtering and structured parsing.}
    \label{fig:DLF}
\end{figure*}

\paragraph{\textbf{Layer 2: GPT-Based Blood Donation Message Classifier}}

We utilize GPT-4o-mini to further filter out non-blood donation-related messages and ensure only relevant positive messages are allowed \cite{26an2024rethinking}. However, this does not introduce any additional cost since this is carried out in the same API call that is used for parsing in the subsequent phase. Figure \ref{fig:DLF} presents the overall architecture of dual-layered flitering. 

\begin{figure}[t]
    \centering
    \includegraphics[width=0.35\textwidth, height=0.22\textheight]{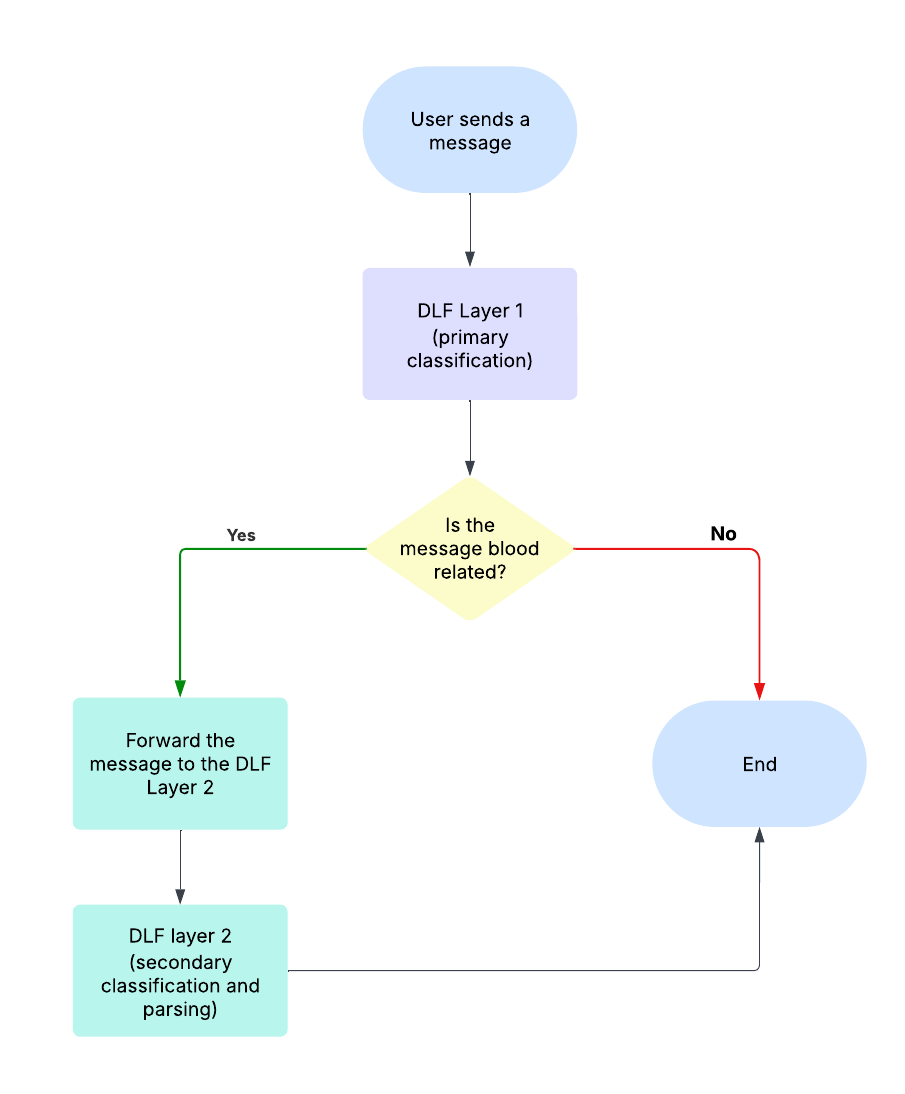}
    \caption{Two-layer DLF framework, with Layer 1 identifying blood-related messages and Layer 2 employing an LLM for detailed filtering and structured parsing.}
    \label{fig:flowchart}
\end{figure}

DLF Layer 1 serves as a lightweight binary classifier that independently filters incoming messages to determine whether they are related to blood request. This layer is optimized for speed and resource efficiency, ensuring that only relevant messages proceed further in the pipeline. DLF Layer 2, powered by an LLM, operates as an independent layer that performs secondary classification to reduce false negatives and conducts detailed parsing on messages identified as blood related by the first layer. By employing this two-tier architecture, we significantly reduce unnecessary API calls to the LLM, thereby optimizing both cost and performance as shown in Figure \ref{fig:flowchart}.

% ANIK
% here, I am removing the contact number from the JSON, since it will not actually matter while deciding which donor to notify next

\subsection{Structured Parsing with Few-shot Prompting}

After detecting a message requesting blood donation, we first parse it into a predefined structure to extract key information. Formally, each message is parsed into a structured object \( M_p \) with fields such as \texttt{blood\_group}, \texttt{bags\_needed}, \texttt{patient}, \texttt{condition}, \texttt{location}, \texttt{hospital\_name}, \texttt{location\_markers}, \texttt{probable\_day}, \texttt{probable\_time}, \texttt{contacts}, and \texttt{compensation} ensuring that all critical elements are captured for efficient processing. For the task of parsing, we finetune the LLama-3.2-3B model using LoRA on a split of our parsing dataset. In case of testing other LLMs, to further enhance parsing precision, we apply the technique of Few-Shot Prompting \cite{27reynolds2021prompt}. In this approach, the model is exposed to a small number of examples, specifically three positive examples and two negative example to guide its predictions. For a positive example message \( M_p \), which is relevant to blood donation, the model is expected to output the parsed information in a structured JSON format. The output can be shown as: $P(M_p) = \text{JSON(properties)}$, where \( P(M_p) \) is the parsed JSON output containing the necessary details for blood donation.
For a negative example message \( M_n \), which is unrelated to blood donation, the model is expected to flag it as irrelevant. The output of the model is: \( P(M_n) = \text{FLAG}_{\text{negative}} \)
where \( P(M_n) \) indicates that the message does not pertain to blood donation.

\begin{align*}
P(M_{\text{new}}) = \begin{cases}
% \text{JSON\{blood\_group, bags\_needed,} \\
% \quad \text{location}\} & \text{if relevant} \\
\text{JSON(properties)} & \text{if relevant} \\
\text{FLAG}_{\text{negative}} & \text{if irrelevant}
\end{cases}
\end{align*}

\section{Experimental Setup}

\subsection{Classifier}

We train and compare among multiple lightweight machine learning classifiers on text embeddings generated using different methods. For BERT-based classifiers such as DistilBERT and MobileBERT, we perform end-to-end training using 3 epochs and a batch size of 2. All embedding generation, training, and evaluation tasks were carried out on a 2$\times$T4 GPU cluster hosted on Kaggle\footnote{\url{https://www.kaggle.com/}}. The models are evaluated based on standard metrics: precision, recall, accuracy, and F1-score. The performance comparison of these first-layer classifiers is reported accordingly.

\subsection{Parser}

To manage inference costs while maintaining evaluation fidelity, we conduct parsing experiments on a stratified random sample of 958 blood donation request texts. This subset includes 329 English, 381 Bengali, and 248 transliterated Bengali messages. First, to generate a gold set of parsed json objects corresponding to the texts, annotations are initially generated using few-shot prompting with the DeepSeek-V3 model. Subsequently, the text-annotation pairs are distributed among five human annotators to ensure robust evaluation. Each sample is assigned to three annotators, who independently assess the correctness and provide a binary verdict (agreement or disagreement). A sample is re-annotated by human annotators if the majority of the assigned annotators disagree with the initial annotation. We evaluate a range of LLMs in both zero-shot and few-shot settings. Parsing accuracy is reported using a weighted score, with 20\% weight on tree edit distance and the rest on field level accuracy. To calculate the tree edit distance, we utilize the \textit{zss} library from Python. For full-precision inference with open-weight models, we utilize the Together AI API\footnote{\url{https://www.together.ai/}} and OpenRouter API\footnote{\url{https://openrouter.ai/}} based on availability. For models from OpenAI, we use the official OpenAI API\footnote{\url{https://platform.openai.com/}}. The LLM decoding parameters for both zero-shot and few-shot inference during parsing were: temparature = 0.7, top\_p = 0.8, top\_k = 35.

We also finetune the LLama-3.2-3B model using LoRA($r = 32, \alpha = 16$) and 4-bit integer quantization, dropout = 0.05, batch size=2, epoch=5, learning rate = $2 \times 10^{-4}$. We use a 80:10:10 split for train, test and validation. 0.81\% of the total 3B parameters are thereby trained on 7.9K paired text and parsed JSON samples. We use a 2$\times$T4 GPU cluster hosted on Kaggle\footnote{\url{https://www.kaggle.com/}} to carry out the finetuning.

\section{Results and Discussion}

\paragraph{DLF Outperforms Other Lightweight classifiers in Accuracy and Efficiency}
We compare DLF with a diverse range of embedding and classifier-based state-of-the-art models for message filtering in Table \ref{tab:DLF}. We experiment with feature extraction methods, including traditional TFIDF \cite{salton1988term} and CountVectorizer (Count) \cite{manning1999foundations}, followed by classifiers such as Logistic Regression (LogReg), Support Vector Machine (SVM) \cite{cortes1995support}, Random Forest (RF) \cite{breiman2001random}, and Naive Bayes (NB) \cite{mccallum1998comparison}. Additionally, we test various pre-trained embeddings with these classifiers, such as Word2Vec (W2V) \cite{mikolov2013efficient}, MiniLM-L6-V2 (MiniLM6) and MiniLM-L12-V2 (MiniLM12), lightweight transformer models for general-purpose sentence embeddings, Paraphrase-MiniLM-L12-v2 (ParaMiniLM) \cite{reimers2019sentence}, DistilUSE \cite{reimers2019sentence}, E5-Small, LaBSE \cite{feng2020language}, Jina Embeddings-V2 (JinaEmb) \cite{jinaai2023}, and BAAI General Embeddings (BGE) \cite{bge-m3}. We also explore end-to-end training of BERT-based classifiers, such as, DistilBERT \cite{Sanh-DistilBERT-2019} and MobileBERT \cite{sun-etal-2020-mobilebert}. As shown in in Table \ref{tab:DLF}, DLF either outperforms or matches the accuracy of other classifiers, while providing the fastest inference.

\begin{table}[h]
\centering
\caption{Comparative performance of filtering methods, where DLF consistently matches or outperforms others across accuracy metrics while achieving the lowest inference time.}
\label{tab:DLF}
\resizebox{\columnwidth}{!}{%
\begin{tabular}{@{}lcccccccc@{}}
\toprule
\textbf{Embedding} & \textbf{Classifier} & \textbf{Accuracy} & \textbf{Precision} & \textbf{Recall} & \textbf{F1-Score} & \makecell[l]{\textbf{Inference}\\\textbf{Time x e-07}\\\textbf{(Seconds)}} \\
\midrule
\multirow{4}{*}{TFIDF} & LogReg & 0.98 & 0.98 & 0.98 & 0.98 &  1.25\\
 & SVM & 0.98 & 0.98 & 0.98 & 0.98 & 2355.85\\
 & RF & 0.98 & 0.98 & 0.98 & 0.98 & 347.45\\
 & NB & 0.97 & 0.97 & 0.97 & 0.97 & 2.06 \\
 \midrule
\multirow{4}{*}{Count} & LogReg & 0.98 & 0.98 & 0.98 & 0.98 & 1.19\\
 & SVM & 0.98 & 0.98 & 0.98 & 0.98 & 1835.88\\
 & RF & 0.98 & 0.98 & 0.98 & 0.98 & 326.89\\
 & NB & 0.96 & 0.96 & 0.96 & 0.96 & 1.82 \\
\multirow{3}{*}{W2V} & LogReg & 0.83 & 0.80 & 0.87 & 0.81 & 10.87\\
 & SVM & 0.82 & 0.80 & 0.86 & 0.81 & 9659.30\\
 & RF & 0.83 & 0.80 & 0.87 & 0.81 & 156.89\\
 \midrule
\multirow{3}{*}{MiniLM6} & LogReg & 0.97 & 0.96 & 0.97 & 0.96 & 6.86\\
 & SVM & 0.97 & 0.97 & 0.97 & 0.97 & 3387.89\\
 & RF & 0.96 & 0.96 & 0.96 & 0.96 & 205.60\\
 \midrule
\multirow{3}{*}{MiniLM12} & LogReg & 0.97 & 0.96 & 0.97 & 0.97 & 6.80 \\
 & SVM & 0.97 & 0.97 & 0.97 & 0.97 & 3028.26 \\
 & RF & 0.97 & 0.96 & 0.97 & 0.96 & 195.58 \\
 \midrule
\multirow{3}{*}{ParaMiniLM} & LogReg & 0.97 & 0.97 & 0.97 & 0.97 & 17.86\\
 & SVM & 0.98 & 0.97 & 0.98 & 0.98 & 2541.56 \\
 & RF & 0.97 & 0.97 & 0.97 & 0.97 & 187.06\\
 \midrule
\multirow{3}{*}{DistilUse} & LogReg & 0.95 & 0.95 & 0.95 & 0.95 & 12.56\\
 & SVM & 0.96 & 0.96 & 0.96 & 0.96 & 6075.95\\
 & RF & 0.97 & 0.96 & 0.97 & 0.97 & 192.09\\
 \midrule
\multirow{3}{*}{E5-Small} & LogReg & 0.98 & 0.98 & 0.98 & 0.98 & 5.94\\
 & SVM & 0.98 & 0.98 & 0.98 & 0.98 & 1933.32\\
 & RF & 0.98 & 0.98 & 0.98 & 0.98 & 186.73\\
 \midrule
\multirow{3}{*}{LaBSE} & LogReg & 0.98 & 0.98 & 0.98 & 0.98 & 11.40\\
 & SVM & 0.98 & 0.98 & 0.98 & 0.98 & 2940.37\\
 & RF & 0.98 & 0.98 & 0.98 & 0.98 & 203.78\\
 \midrule
\multirow{1}{*}{JinaEmb} & LogReg & 0.97 & 0.97 & 0.97 & 0.97 & 19.26\\
\midrule
\multirow{3}{*}{BGE} & LogReg & 0.97 & 0.97 & 0.97 & 0.97 \\
 & SVM & 0.97 & 0.97 & 0.97 & 0.97 \\
 & RF & 0.97 & 0.97 & 0.97 & 0.97 \\

 \midrule
DistilBERT & DistilBERT & 0.98 & 0.98 & 0.98 & 0.98 & 127816.15\\

 \midrule
MobileBERT & MobileBERT & 0.98 & 0.98 & 0.97 & 0.97 & 169240.89\\

 \midrule
\textbf{} & \textbf{DLF} & \textbf{0.99} & \textbf{0.99} & \textbf{0.98} & \textbf{0.98} & \textbf{1.10}\\

\bottomrule
\end{tabular}%
}
\end{table}

\begin{figure}[t]
    \centering
    \includegraphics[width=0.35\textwidth]{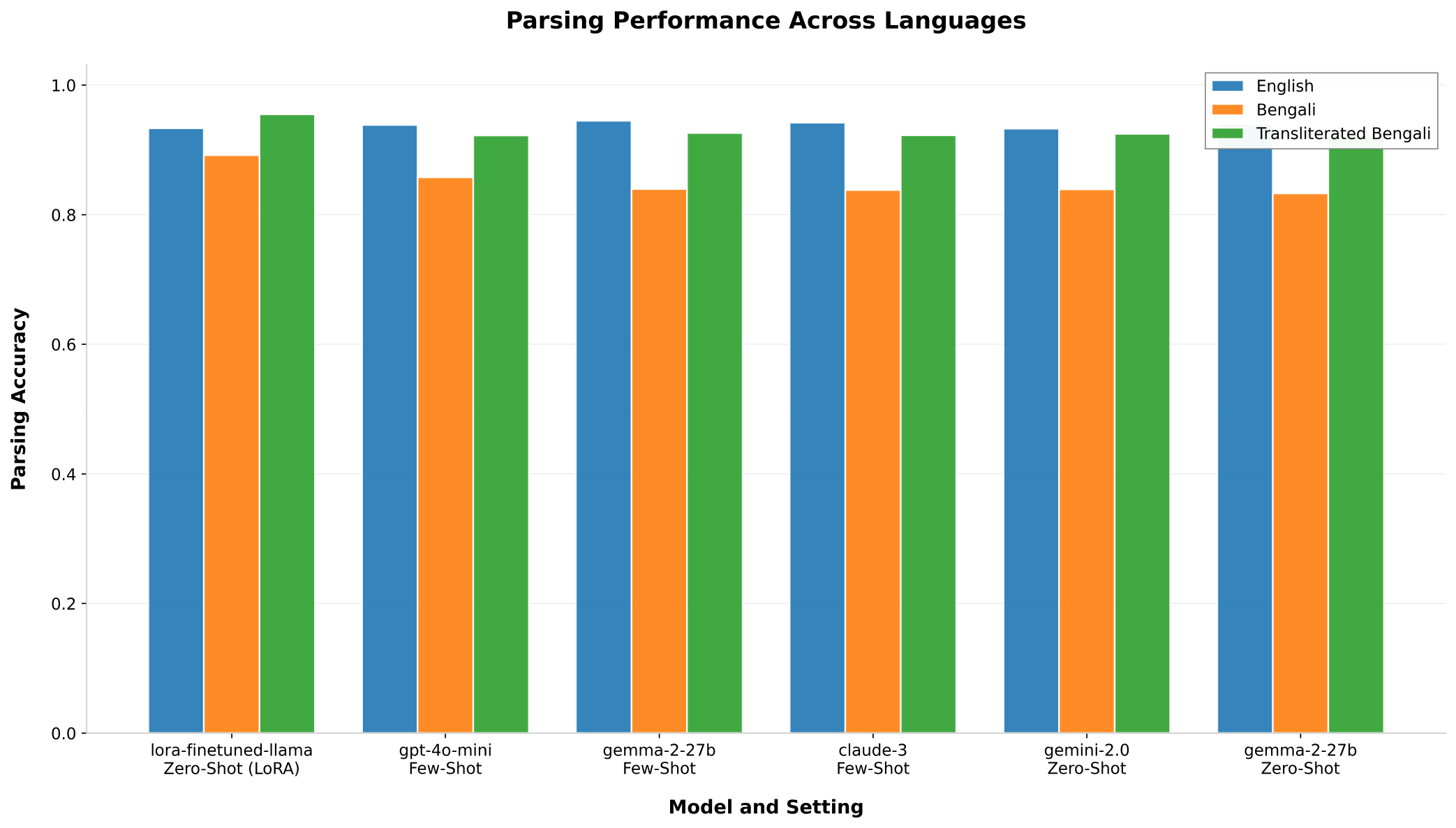}
    \caption{Comparison of parsing accuracy across different languages - Bengali, English, and Transliterated Bengali with the six highest performing model and prompt setting combinations.}
    \label{fig:comparison-of-parsing-accuracy-across-languages}
\end{figure}

\paragraph{LoRA-finetuned Lightweight Parser Outperforms Other Language Models}
Table~\ref{tab:parsing-accuracy} presents the accuracy score under few-shot and zero-shot prompting for various language models, such as, Claude-3-haiku \cite{claude3haiku}, Gemini-2.0 \cite{gemini2}, Gemma-2-27B \cite{gemma2_27b}, GPT-4o-mini \cite{gpt4o_mini}, LLaMA-3.1-8B \cite{meta_llama3_8b}, Meta-LLaMA-3.2-3B \cite{meta_llama3_3b}, LLaMA-3.3-70B \cite{meta_llama3_70b}, Mistral-7B\cite{mistral7b}, Qwen-2.5-7B \cite{qwen2_5_7b}, and our LoRA finetuned LLama-3.2-3B model. While few-shot prompting understandably increases the parsing performance, the finetuned model shows even higher accuracy with zero-shot prompting. Our LoRA finetuned model achieves 92\% zero-shot accuracy surpassing the base model's zero-shot performance by 41.54\% and exceeding the few-shot performance of GPT-4o-mini, Gemini-2.0-flash and other LLMs. Claude-3-haiku, GPT-4o-mini and Gemma-2-27B stand out in few-shot setting with a score of 0.90 each. In case of zero-shot prompting, Gemma-2-27B and GPT-4o-mini also perform strongly, achieving scores of 0.89 and 0.88 respectively. Notably, Mistral-7B shows a severe drop from 0.81 to 0.02 when we shift to zero-shot prompting from few-shot approach, revealing limited generalization. In contrast, Gemma-2-27B, Gemini-2.0, and GPT-4o-mini demonstrate consistent, balanced performance across both settings. Although larger models like LLaMA-3.3-70B outperform the smaller variants as expected, the strong parsing accuracy of the  finetuned model highlights the efficiency of our lightweight model. Figure~\ref{fig:comparison-of-parsing-accuracy-across-languages} presents the parsing accuracy scores across Bengali, English, and Transliterated Bengali for the top six performing model and prompt setting combinations.

\begin{table}[h]
\centering
\caption{Comparison of the parsing accuracy of language models in few-shot and zero-shot settings. Our LoRA-finetuned Llama-3.2-3B model performs strongly even in zero-shot setting surpassing the few-shot performance of other models.
}
\label{tab:parsing-accuracy}
\resizebox{\columnwidth}{!}{%
\begin{tabular}{@{}lll@{}}
\toprule
\textbf{Model} & \textbf{Few-Shot} & \textbf{Zero-Shot} \\ \midrule
Claude-3-haiku & 0.90 & 0.57 \\
Gemini-2.0 & 0.88 & 0.87 \\
Gemma-2-27B & 0.90 & 0.89 \\
GPT-4o-mini & 0.90 & 0.88 \\
LLaMA-3.1-8B & 0.85 & 0.74 \\
LLaMA-3.2-3B & 0.68 & 0.65 \\
LLaMA-3.3-70B & 0.88 & 0.87 \\
Mistral-7B & 0.81 & 0.02 \\
Qwen-2.5-7B & 0.83 & 0.78 \\
LoRA-finetuned LLama-3.2-3B & - & \textbf{0.92} \\
% \textbf{DeepSeek-V3}  & \textbf{1.00} & \textbf{0.88} \\
\bottomrule
\end{tabular}%
}
\end{table}

\paragraph{LoRA-finetuned Lightweight Parser Enables Time-efficient and Token-efficient Parsing}

We notice that both TFIDF and Count embeddings show the fastest inference time when paired with LogReg and NB classifiers in Table \ref{tab:DLF}. LogReg and NB consistently show the lowest inference time across most embeddings. We compare complexity across different models using the CBRS dataset and evaluate them separately for Bengali, English, and Transliterated Bengali, as well as for the total dataset in Table \ref{tab:computational}. In this case, our finetuned 3B parameter model significantly reduces the input token usage since we only need to pass the message without any instruction or examples in this case. On average, it only requires 78.50 input tokens, in contrast to almost 3K average input tokens for other LLMs. We find that Gemini-2.0 and LLaMA-3.2-3B offer the lowest average cost. We observe that Gemma-2-27B and LLaMA-3.3-70B have the longest inference time. In contrast, LLaMA-3.1-8B, LLaMA-3.2-3B, and Qwen-2.5-7B perform significantly faster, with inference time ranging from 1.02 to 2.27 seconds. We note that most models exhibit consistent performance across languages, with slightly higher computational cost for Bengali. It suggests strong robustness across linguistic contexts. However, Gemma-2-27B shows the highest inference time of 5.42 seconds for Bengali which indicates that Bengali may require more computational effort due to linguistic complexity, a larger vocabulary, or less optimized processing for Bengali. We also find that English and Transliterated Bengali generally have lower and more consistent inference times. For most models, the average cost varies only slightly across languages. Some models show a marginal cost increase for Bengali. Claude-3-Haiku costs 0.00107 for bn, compared to 0.00095 for en and 0.00094 for tbn. Similarly, LLaMA-3.2-3B costs 0.00020 for bn, slightly higher than 0.00017 for en and 0.00017 for tbn due to its longer inference time for Bengali.

\begin{table}[h]
\centering
\caption{Comparison of message parsing complexity across LLMs in terms of cost, token count, and inference time, showing that our LoRA-finetuned parser achieves superior efficiency on all metrics.}
\label{tab:computational}
\resizebox{\columnwidth}{!}{%

\begin{tabular}{l l l l l l l}
\toprule
\makecell[l]{\textbf{Model}} 
& \makecell[l]{\textbf{Data}} 
& \makecell[l]{\textbf{Avg}\\\textbf{Cost}} 
& \makecell[l]{\textbf{Avg}\\\textbf{Input}\\\textbf{Tokens}} 
& \makecell[l]{\textbf{Avg}\\\textbf{Output}\\\textbf{Tokens}} 
& \makecell[l]{\textbf{Avg}\\\textbf{Total}\\\textbf{Tokens}} 
& \makecell[l]{\textbf{Inference}\\\textbf{Time}\\\textbf{(Seconds)}} \\
\midrule
\multirow{4}{*}{Claude-3-Haiku} & bn                             & 0.00107           & 2991.29                   & 258.67                    & 3249.96                  & 2.89                       \\
                                & en                             & 0.00095           & 2825.26                   & 198.11                    & 3023.37                  & 2.49                       \\
                                & tbn                            & 0.00094           & 2818.61                   & 191.32                    & 3009.93                  & 2.49                       \\
                                & total                          & 0.00100           & 2889.91                   & 220.56                    & 3110.48                  & 2.65                       \\
\midrule
% \multirow{4}{*}{DeepSeek-V3}    & bn                             & 0.00119           & 2368.67                   & 256.31                    & 2624.98                  & 8.28                       \\
%                                 & en                             & 0.00100           & 2291.03                   & 127.44                    & 2418.47                  & 4.65                       \\
%                                 & tbn                            & 0.00094           & 2285.85                   & 116.14                    & 2401.99                  & 4.25                       \\
%                                 & total                          & 0.00106           & 2320.73                   & 176.03                    & 2496.76                  & 6.04                       \\
% \midrule
\multirow{4}{*}{Gemini-2.0}     & bn                             & 0.00044           & 2845.10                   & 388.46                    & 3233.56                  & 3.87                       \\
                                & en                             & 0.00033           & 2744.49                   & 137.38                    & 2881.87                  & 2.59                       \\
                                & tbn                            & 0.00033           & 2740.46                   & 131.37                    & 2871.83                  & 2.50                       \\
                                & total                          & 0.00037           & 2783.67                   & 236.20                    & 3019.87                  & 3.07                       \\
\midrule
\multirow{4}{*}{Gemma-2-27B}    & bn                             & 0.00251           & 2857.10                   & 279.11                    & 3136.21                  & 5.42                       \\
                                & en                             & 0.00232           & 2756.49                   & 137.57                    & 2894.06                  & 3.32                       \\
                                & tbn                            & 0.00230           & 2752.46                   & 127.94                    & 2880.40                  & 2.95                       \\
                                & total                          & 0.00239           & 2795.67                   & 191.66                    & 2987.33                  & 4.06                       \\
\midrule

\multirow{4}{*}{GPT-4o-mini}    & bn                             & 0.00042           & 2284.57                   & 147.36                    & 2431.93                  & 3.90                       \\
                                & en                             & 0.00040           & 2221.49                   & 114.65                    & 2336.14                  & 3.23                       \\
                                & tbn                            & 0.00040           & 2216.08                   & 106.26                    & 2322.33                  & 3.11                       \\
                                & total                          & 0.00041           & 2245.31                   & 125.55                    & 2370.86                  & 3.47                       \\
\midrule
\multirow{4}{*}{LLaMA-3.2-3B} & bn                             & 0.00020           & 2919.22                   & 315.24                    & 3234.46                  & 3.36                       \\
                                   & en                             & 0.00017           & 2683.68                   & 68.14                     & 2751.83                  & 1.49                       \\
                                   & tbn                            & 0.00017           & 2663.64                   & 93.97                     & 2757.61                  & 1.65                       \\
                                   & total                          & 0.00019           & 2772.65                   & 173.62                    & 2946.27                  & 2.27                       \\
\midrule
\multirow{4}{*}{LLaMA-3.1-8B} & bn                             & 0.00063           & 3040.99                   & 481.06                    & 3522.05                  & 2.41                       \\
                                   & en                             & 0.00053           & 2810.61                   & 120.55                    & 2931.16                  & 1.19                       \\
                                   & tbn                            & 0.00053           & 2805.85                   & 130.25                    & 2936.09                  & 1.02                       \\
                                   & total                          & 0.00057           & 2901.48                   & 267.19                    & 3168.67                  & 1.63                       \\
\midrule

\multirow{4}{*}{LLaMA-3.3-70B} & bn                             & 0.00288           & 3042.21                   & 236.08                    & 3278.28                  & 4.86                       \\
                                    & en                             & 0.00258           & 2812.09                   & 121.07                    & 2933.17                  & 3.54                       \\
                                    & tbn                            & 0.00257           & 2807.41                   & 110.11                    & 2917.52                  & 3.28                       \\
                                    & total                          & 0.00270           & 2902.88                   & 164.21                    & 3067.09                  & 4.00                       \\
\midrule
\multirow{4}{*}{Mistral-7B}         & bn                             & 0.00075           & 3524.80                   & 215.26                    & 3740.06                  & 2.67                       \\
                                    & en                             & 0.00070           & 3313.11                   & 163.78                    & 3476.89                  & 2.20                       \\
                                    & tbn                            & 0.00069           & 3307.23                   & 163.34                    & 3470.57                  & 2.24                       \\
                                    & total                          & 0.00072           & 3396.22                   & 184.25                    & 3580.46                  & 2.40                       \\
\midrule
\multirow{4}{*}{Qwen-2.5-7B}        & bn                             & 0.00100           & 3113.89                   & 212.39                    & 3326.28                  & 2.27                       \\
                                    & en                             & 0.00092           & 2913.79                   & 139.69                    & 3053.49                  & 1.65                       \\
                                    & tbn                            & 0.00091           & 2908.61                   & 125.06                    & 3033.67                  & 1.57                       \\
                                    & total                          & 0.00095           & 2992.45                   & 164.97                    & 3157.41                  & 1.88                       \\

\midrule
\multirow{1}{*}{LoRA finetuned LLama-3.2-3B} 
                                    & total                          & -           & \textbf{78.50}                  & \textbf{189.08 }                   & \textbf{267.58}                  & 1.35                       \\

\bottomrule
\end{tabular}
}
\end{table}

\section{Error Analysis}

Through manual inspection of the predictions from both the classifier and parsing models, we conducted a qualitative error analysis. Below, we summarize the key sources of errors.

\subsection{Classifier Errors}

\paragraph{Appreciation Messages.}  
Public posts on Facebook often contain appreciation messages such as, "We are very grateful to X for donating A+ blood at location Y." These posts are not actual blood donation requests, yet the first-layer classifier occasionally misclassifies them as such.

\paragraph{Edited Messages.}  
Messages on Telegram can be edited after posting, which introduces another source of error. For example, a message such as, "Update: Managed, Emergency blood needed, \ldots" may have originally been a blood donation request but was later marked as already managed. Such cases also tend to be misclassified by the first layer.

\subsection{Parser Errors}

\paragraph{Distorted Locations.}  
Locations mentioned in the messages are often lengthy or appear in disjoint segments, making them difficult to parse accurately. Although the models were instructed to preserve the original language and structure of the location, they frequently altered it. This issue contributed to the significant drop in zero-shot performance of Mistral-7B, Claude-3-Haiku, and Llama-3.1-8B (Table~\ref{tab:parsing-accuracy}).

\paragraph{Structural Deformation.}  
Several models, including Mistral-7B and Claude-3-Haiku, struggled to adhere to the required JSON structure in zero-shot settings. Common issues included dropping fields, introducing unwanted fields, or modifying the expected field names.

\paragraph{Bengali Messages.}  
Parsing Bengali messages posed particular challenges, as also reflected in Figure~\ref{fig:comparison-of-parsing-accuracy-across-languages}. Errors included misinterpretation of blood groups written in Bengali, distortion of time and location expressions, and failure to identify valid fields.

\section{Conclusion}
In our study, we present CBRS-a system combining a curated dataset with a multi-platform bot for social media groups. Due to the lack of low-resource language datasets and the informal nature of online communication, extracting relevant information from large message streams is challenging. We propose a dual-layer filtering and parsing architecture for efficient extraction from Bengali, English, and Transliterated Bengali. 
This advances object-based filtering in task-specific domains and lays the groundwork for intelligent, cross-platform bots in healthcare.

\section{Ethical Considerations}

This study was conducted in accordance with institutional ethical guidelines. The collection of data from publicly accessible social media communities on Facebook and Telegram was approved by the Institutional Ethics Review Board. To protect privacy, all identifying information of the source users from collected posts and messages was anonymized. In addition, we ensure that, even though the messages and posts in the blood request messages contain patient's health information, they do not contain patient's name or any personal identifier. During the study, all survey participants provided informed consent, and all personal identifiers were removed prior to analysis.

\section{Limitations}
The current study focuses only on Bengali and English, limiting broader multilingual applicability and cross-regional validation. Future work will expand to more languages and regions to improve generalizability. Although CBRS performs well, it faces challenges in scalability, message storage, and spam control, which could overwhelm donors with irrelevant requests. Future improvements will optimize storage, address spam risks, evaluate performance at larger scales, and expand to more platforms to increase accessibility and usability.

\bibliography{references}

\appendix

\begin{figure}[t]
    \centering
    \includegraphics[width=0.4\textwidth]{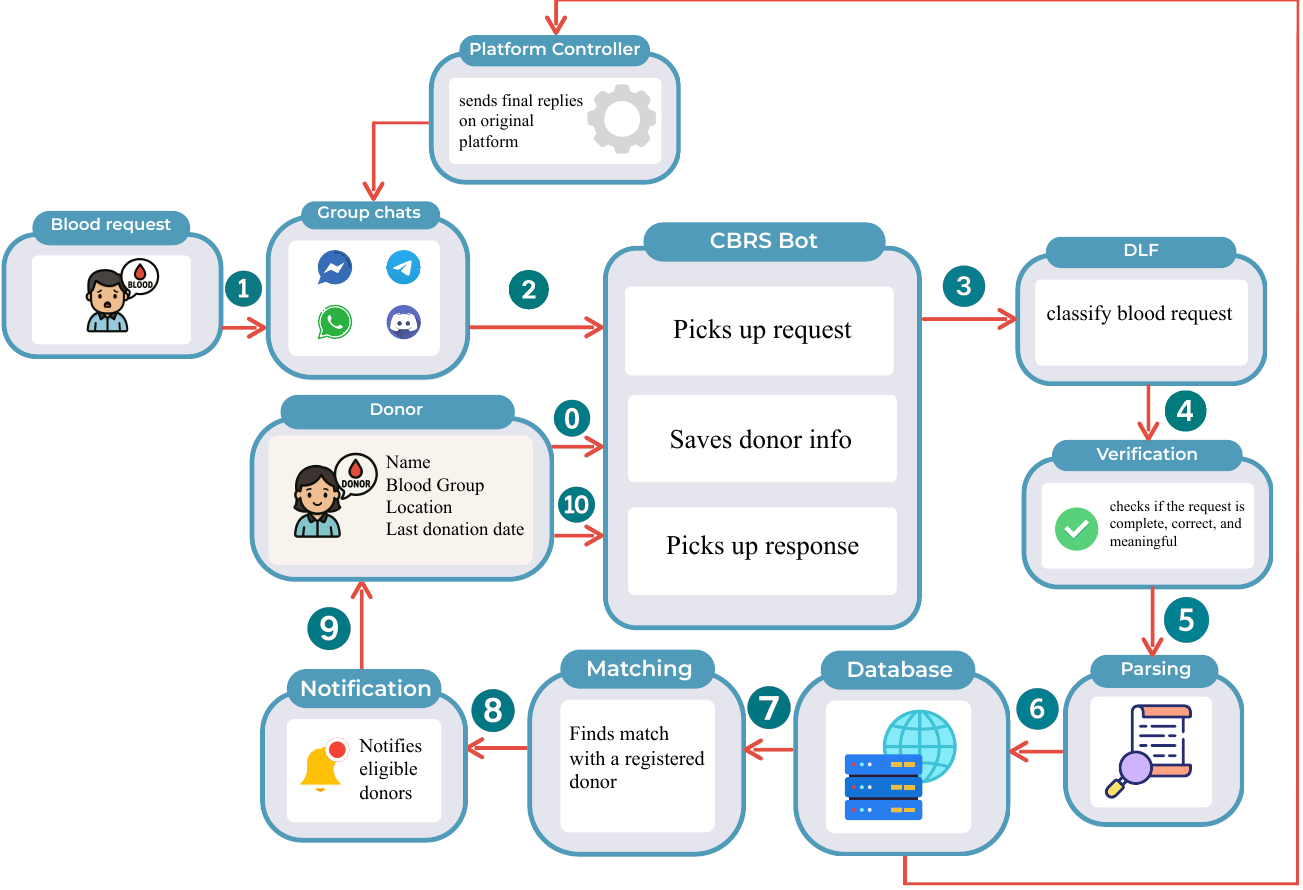}
    \caption{This figure illustrates the overall workflow of CBRS. After filtering and parsing through DLF, a notification is sent to potential donors. Additionally, it outlines an integrated strategy for seamless donor engagement.}
    \label{fig:archi}
\end{figure}

\section{Prompts}
The details architecture of the system is shown in Figure \ref{fig:archi}.
The prompt used for parsing the free-form text messages into structured JSON objects is given in Figure \ref{fig:fewshot-prompt-template}. The prompt used to curate adversarial negative samples is given in Figure \ref{fig:fewshot-adversarial-template}. 

\begin{figure*}[t]
\centering
\begin{tcolorbox}[title=\textbf{Few-shot Prompt for Blood Donation Request Parsing}, width=0.95\textwidth, colback=white]
\textbf{SYSTEM:} \\
You will be provided with a message, typically sent by an individual or organization, which may pertain to a request for blood donation. Your task is to determine whether the message is a blood donation request, and if yes, then to extract the necessary information. 

\textbf{Instructions:}
\begin{itemize}
    \item If the message is \textbf{not} a blood donation request, respond with: \verb|{"is_blood_donation_request": false}|. No other fields are required.
    \item If it \textbf{is} a request, extract relevant information into a well-structured JSON object strictly conforming to the schema.
    \item Set fields to \verb|""| if not stated explicitly in the message.
\end{itemize}

\textbf{Schema:}
\begin{itemize}
    \item \texttt{blood\_group}: one of [A+, A-, B+, B-, O+, O-, AB+, AB-] or \verb|""|
    \item \texttt{bags\_needed}: string (e.g., \verb|"3"| or \verb|"3-4"|)
    \item \texttt{patient}: \{name, gender [M/F/""], age\_group [child/teenager/young/adult/""]
    \item \texttt{condition}: comma-separated medical conditions or status
    \item \texttt{location}, \texttt{hospital\_name}: as stated
    \item \texttt{location\_markers}: list of city/region tokens
    \item \texttt{probable\_day}: one of [\texttt{DD/MM}, \texttt{DD/MM/YYYY}, \texttt{today}, \texttt{tomorrow}, \texttt{n days later}]
    \item \texttt{probable\_time}: one of [\texttt{HH:MM}, \texttt{before HH:MM}, \texttt{after HH:MM}, \texttt{HH:MM-HH:MM}, \texttt{in n hours}] (24-hr format)
    \item \texttt{contacts}: list of \{name, contact\_numbers [\ldots], relation\_with\_patient\}
    \item \texttt{compensation}: \{transportation: [Y/N/""], allowance: [Y/N/""]\}
\end{itemize}

\vspace{0.3em}
\textbf{Examples:}
\begin{itemize}
    \item \textbf{\texttt{<positive\_example>}}
    \item \textbf{\texttt{<negative\_example>}}
\end{itemize}

\vspace{0.3em}
\textbf{Final Query:} \\
\textit{Text Message:} \texttt{\{user\_text\}} \\
\textbf{Instruction:} Output only the valid JSON response. No explanations, greetings, or hallucinations.
\end{tcolorbox}
\caption{Few-shot prompt for blood donation request parsing.}
\label{fig:fewshot-prompt-template}
\end{figure*}

\begin{figure*}[t]
\centering
\begin{tcolorbox}[title=\textbf{Few-shot Prompt Template for Adversarial Negative Sample Generation}, width=0.95\textwidth, colback=white]
\textbf{SYSTEM:} \\
You are tasked with generating adversarial examples for a text classification model designed to identify blood donation-seeking messages. The goal is to create realistic, diverse, and tricky negative examples that are \textbf{not} actual blood donation requests, but use vocabulary commonly associated with such requests.

\textbf{Vocabulary:}
\begin{itemize}
    \item \textbf{Bengali Words:} \texttt{<bengali\_words>}
    \item \textbf{English Words:} \texttt{<english\_words>}
\end{itemize}

\textbf{Anchor Examples:} \\
Use the provided \texttt{<negative\_examples>} as inspiration. You must create similarly styled yet novel adversarial examples that closely mimic the linguistic pattern of real blood donation messages.

\textbf{Output Format:} \\
Generate \texttt{\{num\_examples\}} new examples in JSON format. Each example must be a JSON object with the following keys:
\begin{itemize}
    \item \texttt{"en"}: The message in English
    \item \texttt{"bn"}: The equivalent Bengali translation
    \item \texttt{"tbn"}: The transliterated Bengali text in Latin script
\end{itemize}
Return the output as a single JSON array of these objects. All strings must be enclosed in double quotes and follow correct JSON syntax.

\textbf{Guidelines:}
\begin{itemize}
    \item \textbf{Realistic and Diverse}: Make the messages resemble real-world posts (e.g., social media, chat, awareness campaigns) but ensure they are not genuine blood donation requests.
    \item \textbf{Tricky}: Use several keywords from the provided lists in each example to make it deceptively similar to an actual request.
    \item \textbf{Not Genuine Requests}: The core content must not represent a legitimate need for blood donation.
\end{itemize}

\textbf{Reminders:}
\begin{itemize}
    \item Strictly follow the JSON schema.
    \item Do \textbf{not} include any greetings, comments, or explanations in the output.
    \item All examples must be \textbf{distinct}, \textbf{novel}, and \textbf{realistic}.
\end{itemize}

\vspace{0.3em}
\textbf{Final Query:} \\
Generate <num\_examples> adversarial examples in the specified JSON format.
\end{tcolorbox}
\caption{Few-shot prompt for generating adversarial negative samples for blood donation message classification.}
\label{fig:fewshot-adversarial-template}
\end{figure*}

 \section{Detailed Workflow of CBRS} \label{workflow}
 \paragraph{System Integration Flow}
The bots are initially integrated into social groups with the explicit consent of both users and administrators. The bot serves two purposes. Firstly, it encourages group members to register as donors by providing a direct link to the registration inbox. Secondly, it polls for messages in the group continuously and looks for ones that seek blood donations.

\paragraph{Donor Enrollment Process}
When users engage with our bot via direct messaging to register as donors, they are redirected to a centralized registration web application. This interface systematically collects a data set \( D = \{ \text{blood\_group}, \text{current\_location}, \text{last\_donation\_date} \} \). We use browser geo-location to get accurate latitude and longitude, with explicit user consent before data collection. Upon submission, the dataset \( D \) is stored with the corresponding chat platform \( ID \), allowing efficient user notifications for future blood donation requests. The interface allows users to update their information at any time. This ensures accurate tracking of their last donation date for better record maintenance. The donor enrollment is a single-point of design choice made to streamline input across existing multiple platform interfaces.
 \paragraph{User Interface}
The CBRS interface majorly comprises two components: the chatbot interface and a single point of donor information intake. Since we plan to employ our bots as members of already running chat groups, the chat interface is essentially the same as the interfaces of those corresponding chat platforms. For our initial design, we selected two prominent chat platforms named Telegram and Discord. Both of these platforms feature engaging conversational interfaces which we utilize for our purpose. Unlike several other platforms, Telegram and Discord share a particular feature, namely, the use of slash user commands in the chat interface. Since the interactions with our chatbot are limited in possible options, we opt to design convenient user commands rather than parsing natural language messages from users on the fly shown in Table \ref{tab:bot_commands} . The currently available user commands in our chat interface are as follows:

\begin{table}[h]
    \centering
    \caption{Bot Commands and their Purposes}
    \label{tab:bot_commands}
    \resizebox{\columnwidth}{!}{%
    \begin{tabular}{@{}ll@{}}
        \toprule
        \textbf{Command} & \textbf{Purpose} \\
        \midrule
        \texttt{/start}           & Initialize interaction with the bot \\
        \texttt{/help}            & Display a user guide \\
        \texttt{/show\_my\_info}  & Show the registered user details \\
        \texttt{/update\_my\_info}& Update user information \\
        \texttt{/register\_as\_donor} & Register as a blood donor \\
        \texttt{/goodbye}         & End interaction with the bot \\
        \bottomrule
    \end{tabular}%
    }
\end{table}
To facilitate the input of donor information, we designed a single-page web application featuring a form that receives the blood group, last donation date, and GPS location from the browser. To eliminate the need for reiterating a donor’s chat platform identity, we generate a unique URL for the donor based on their user account on the chat platform. By visiting this unique URL, the donor can update their information directly from the chat interface at any time.
\paragraph{Context-Aware Notification Strategy}
Efficient donor notification introduces non-trivial challenges. Firstly, over- or under-notification can impair both user experience and system efficiency. To mitigate this, we adopt an iterative, stage-wise notification strategy, where donors are queried sequentially. Upon receiving a positive response, further alerts are suppressed, and the seeker is immediately informed. The stage-depth is dynamically governed by the urgency level inferred via the parsing model. Secondly, post-hoc message edits-particularly those indicating successful blood acquisition-necessitate retroactive updates. We maintain a notification ledger for each request; upon detecting such edits, prior recipients are promptly notified of the resolution.
\paragraph{Implementation Details}
The chatbots are implemented with standard libraries released and maintained by
the corresponding chat platform. For instance, to design the chatbot for Telegram,
we use the library python-telegram-bot in python and discord.js library for the
Discord bot. These libraries help us take appropriate actions based on slash
commands and user messages. In case of slash commands, we perform string
matching and execute corresponding methods. On the other hand, for any non-command text, we first call the filtering API with the text to determine whether it
is actually seeking blood donation or not. If yes, we further call our parsing API to
parse the text into a JSON format. We first perform training on a curated dataset and then carry out
inferences. We train the model for 1000 epochs using a learning rate 1.0. We use trigrams (wordNgrams=3) to capture better context from word sequences. Subword length is configured with minn=3 and maxn=6 to handle out-of-vocabulary words. The parsing API is implemented with \href{https://www.langchain.com/}{Langchain}. As LLM, we use GPT-4o-mini with few shot prompting. 
The unified donor information intake application is built with \href{https://react.dev/}{React}
 and these
pieces of information are stored in \href{https://www.mongodb.com/products/platform/atlas-database}{MongoDB} under appropriate models

\begin{figure}[!h]
    \centering
    \includegraphics[width=\linewidth]{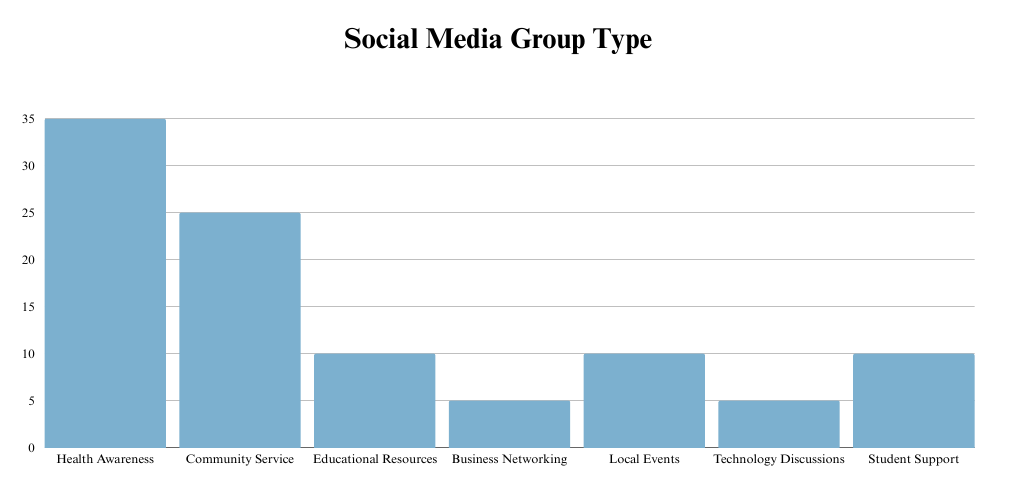}
    \caption{This figure shows demographic distribution by group type.}
    \label{fig:Group_type}
\end{figure}

\begin{figure}[!h]
    \centering
    \includegraphics[width=\linewidth]{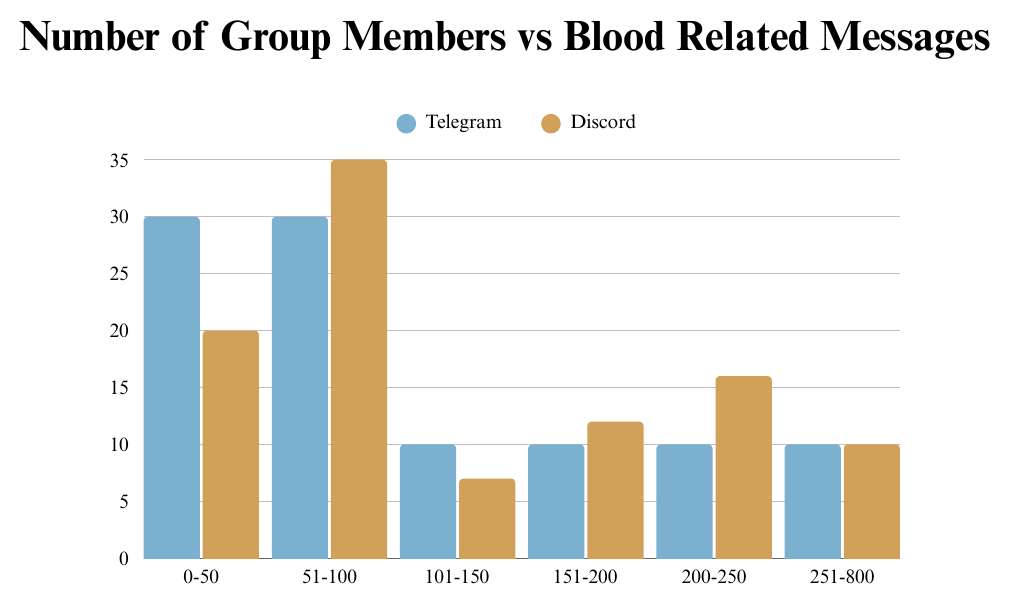}
    \caption{This figure shows demographic distribution by group size.}
    \label{fig:Group_member}
\end{figure}

\begin{figure}[!h]
    \centering
    \includegraphics[width=\linewidth]{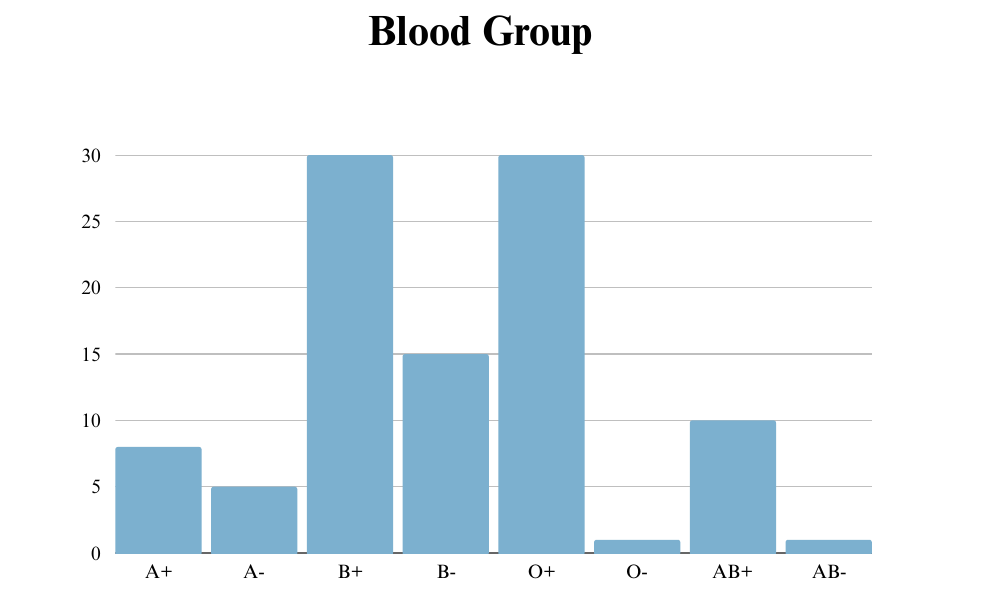}
    \caption{This figure shows blood donation related messages numbers in different groups and blood group of users}
    \label{fig:BloodGroup}
\end{figure}

\begin{figure}
    \centering
    \includegraphics[width=0.8\linewidth]{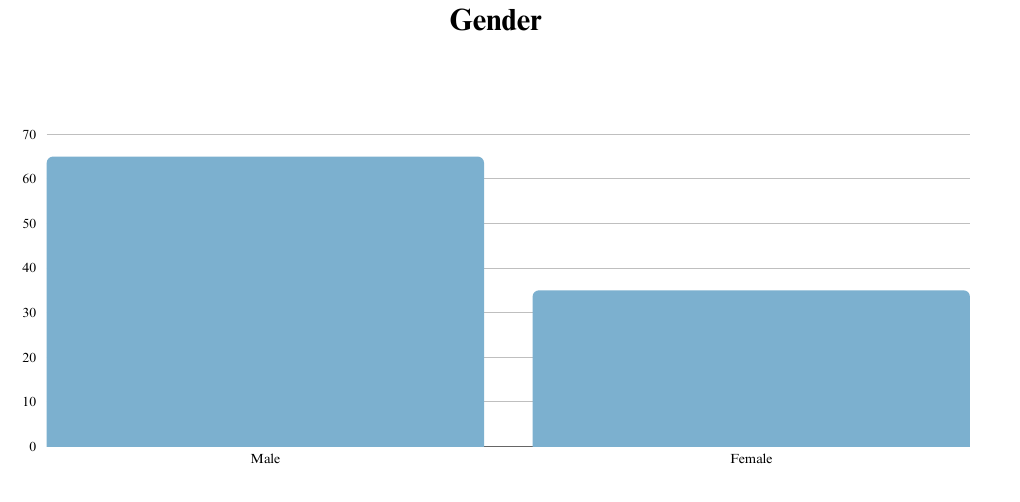}
    \caption{This figure shows demographic distribution by gender.}
    \label{fig:Gender}
\end{figure}
\begin{figure}
    \centering
    \includegraphics[width=0.8\linewidth]{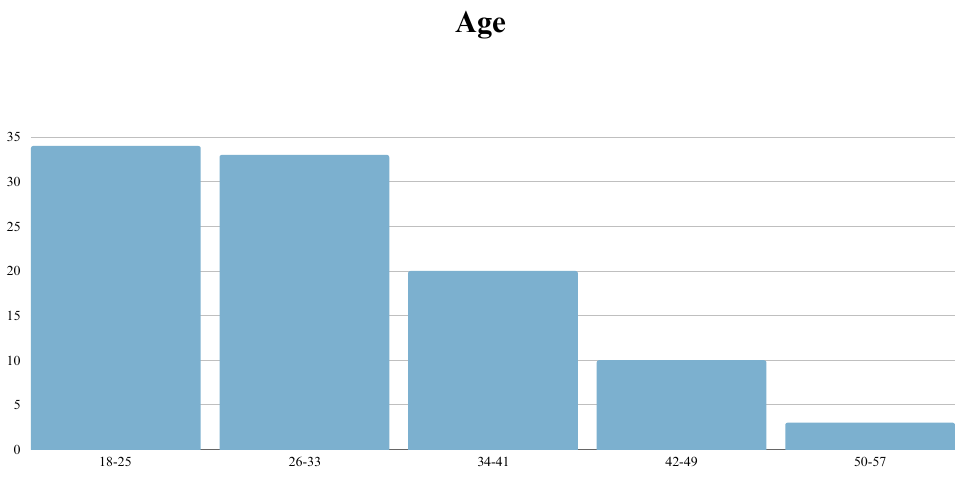}
    \caption{This figure shows demographic distribution by age.}
    \label{fig:Age}
\end{figure}
\begin{figure}
    \centering
    \includegraphics[width=\linewidth]{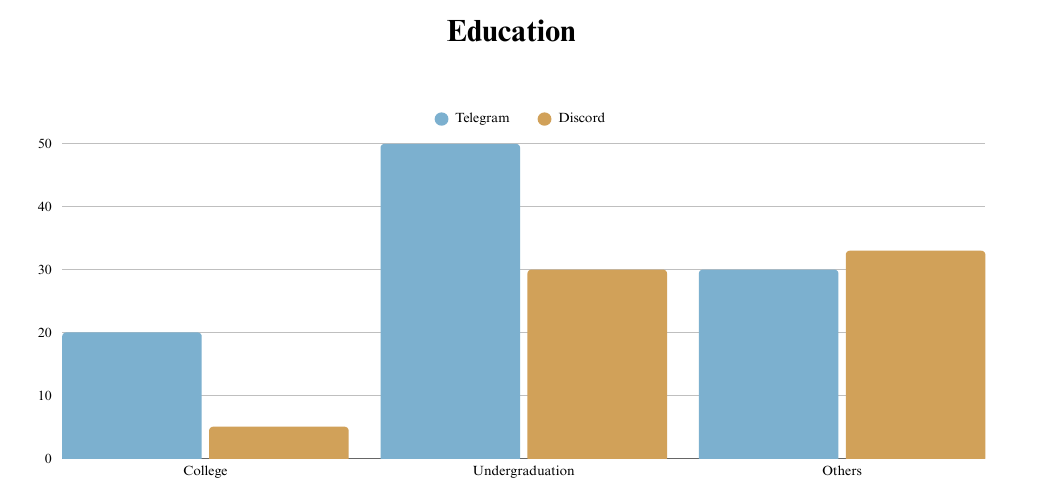}
    \caption{This figure shows demographic distribution by education.}
    \label{fig:Education}
\end{figure}

\begin{figure}
    \centering
    \includegraphics[width=0.8\linewidth]{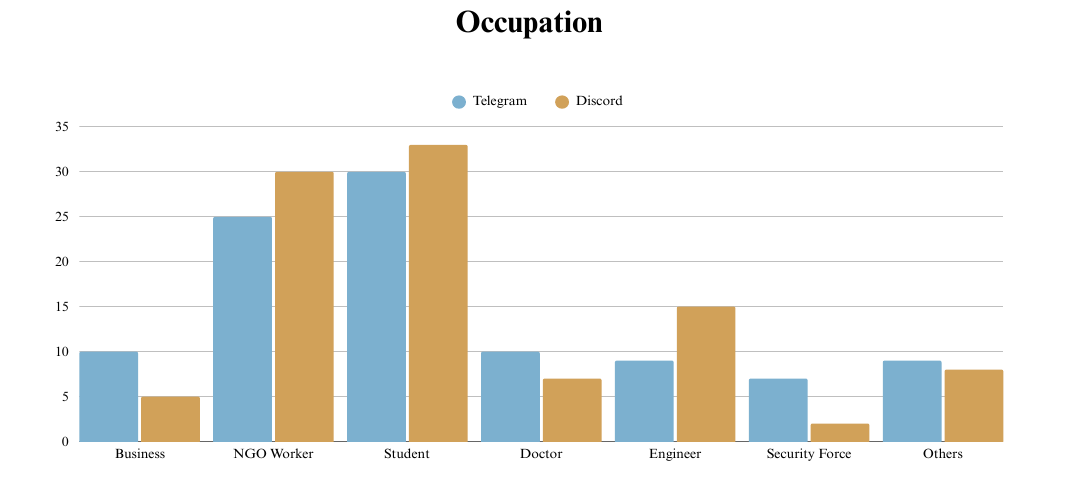}
    \caption{This figure shows demographic distribution by Occupation.}
    \label{fig:Occupation}
\end{figure}
\section{User Study}\label{appendix:B}
To assess the effectiveness of multi-platform bots for timely message filtering and notifications, we conducted a study on various indicators like response time, user interface, command usability and satisfaction levels. Participants shared their experiences regarding delays in receiving blood donation requests. We organized questions into two parts: one for those who frequently share these messages and another for blood donors. Different demographics were included to ensure a balanced study.
\subsection{Conditions}
We conducted a between-subject study with two conditions for a predefined period — the baseline social media group without a bot and one integrated with CBRS. Each condition featured a consistent set of questions designed to gather insights on user experience, response time, engagement levels, and challenges. The baseline system relied on manual messaging and user coordination; in contrast, the CBRS integration introduced blood donation message filtering, automated responses, real-time donor matching, and geo-location-based notifications. 
\subsection{Participants}
We recruited members from 20 active groups on Telegram and 10 active groups on Discord, all based in Bengalidesh. Figure \ref{fig:Group_type} and Figure \ref{fig:Group_member} shows the demographic distribution across the groups. The groups varied in size, ranging from 50 to 8,00 members, and were created for different purposes, including health awareness (35\%), community service (25\%), educational resources (10\%), business networking (5\%), local events (10\%), technology discussions (5\%) and student support (10\%). Furthermore, the volume of messages within these groups ranged from 30 to 110 with 1-20 as blood donation requests per day. A total of 114 participants, with 38 potential donors, joined the survey on pre- and post-integration of CBRS. Figure \ref{fig:BloodGroup}, \ref{fig:Gender}, \ref{fig:Age}, \ref{fig:Education}, and \ref{fig:Occupation} shows the demographic breakdown of a diverse group of participants, aged 18 to 57. Most were between 18-25 (34\%) and 26-33 (33\%). A majority (80\%, 91/114) had at least a college or bachelor’s degree. The participants came from various professions: 10\% in business, 25\% as NGO workers, 10\% as doctors, 9\% as engineers, 7\% in security forces, and 30\% were students.
\subsection{Procedures}
At the start of the study, we selected diverse groups based on different targets and ages. Admins signed consent form from the group. We recorded the average daily messages and blood-related messages per group. Each participant signed a consent form and completed a pre-study questionnaire to gather demographic information including gender, age, occupation, education level, prior experience with social media groups, and other BDSs regarding blood donation initiatives. They also shared their initial expectations from CBRS.

The study was conducted over three days, from October 23 to 26, 2024. We integrated bots into the groups. A total of 108 individuals registered as donors from different locations. The last donation dates and blood groups are stored during registration on bots. Among the donors, 30\% are O+ and 30\% are B+ blood types, while 1\% are O- and AB-, indicating a lower proportion of negative blood types as shown in Figure \ref{fig:BloodGroup}. They did not receive training on the bot to assess the intuitiveness of user interface. We then began collecting feedback from them. Participants were asked about their satisfaction levels in areas such as prior problems, satisfaction with slash-command prompts, user interface, overall functionality, comparisons with other apps, challenges in using the bots, and suggested improvements. Responses were gathered using a five-point Likert scale to evaluate their experiences. Our contributor survey had two types: users who made requests and donors who were notified through bots and donated in the last three days. Users answered 11 questions, while donors answered 7 questions given in Appendix \ref{appendix:C}.

\section{Survey Questionnaire}\label{appendix:C}
We surveyed 114 participants, including 38 potential donors and gathered valuable insights on their satisfaction levels and open feedback regarding challenges and suggestions for improvement. This provided valuable contributions to our work. The survey questions are given below:
\subsection*{For Users:}
\begin{enumerate}
    \item Do you request blood donations on social media (e.g., Telegram, Discord, etc.)?\\
    (Almost always, Often, Sometimes, Seldom, Never)
    
    \item Did you usually receive timely responses to your blood donation requests before using BNet prior to October 23, 2024?\\
    (Almost always, Often, Sometimes, Seldom, Never)

\item How satisfied are you with the timely response of BNet in identifying potential donors between October 23 and October 26, 2024, after integrating BNet into groups?\\(Very satisfied, Satisfied, Neither, Dissatisfied, Very dissatisfied)

      \item After getting a response from BNet, have you successfully connected with a blood donor through BNet?\\
    (Almost always, Often, Sometimes, Seldom, Never)

    \item How easy do you find using BNet through slash command prompts?\\
    (Extremely easy, Very easy, Moderately easy, Slightly easy, Not at all)
    \item How intuitive is the user interface of BNet?\\
    (Extremely intuitive, Very intuitive, Moderately intuitive, Slightly intuitive, Not at all)
    \item How would you rate the overall functionality of BNet?\\
    (Excellent, Above Average, Average, Below Average, Very Poor)
    \item At most how many blood donation seeking messages do you feel comfortable to receive from BNet per month?\\
    (1-5, 6-10, 11-15, 16-20, 21+)
    \item Do you find BNet more effective than existing blood donation apps or methods you have used before?\\
    (Much better, Somewhat better, Stayed the same, Somewhat worse, Much worse, Not applicable- I have never used any app before)

    \item What challenges do you face in connecting with blood donors? How can these be overcome?\\
    (Open-ended response)
    \item  What improvements would you suggest to make BNet better for requesters?\\
    (Open-ended response)
\end{enumerate}
\subsection*{For Donors:}
\begin{enumerate}
    \item How many times have you donated blood in the past year? \\
    (Never, 1 time, 2 times, 3 times, 4 or more)

    \item Do you have trouble finding blood donation requests among a large volume of messages in social media groups? \\
    (Almost always, Often, Sometimes, Seldom, Never)

    \item How convenient is BNet in notifying you about blood donation requests in social media groups?\\
    (Extremely convenient, Very convenient, Moderately convenient, Slightly convenient, Not at all)

     \item How would you rate the overall functionality of BNet? \\
    (Excellent, Above Average, Average, Below Average, Very Poor)

     \item Do you find BNet more effective than existing blood donation apps or methods you’ve used before?\\
    (Much better, Somewhat better, Stayed the same, Somewhat worse, Much worse, Not applicable)

   \item What challenges do you face in connecting with blood requesters? How can these be overcome?\\
 (Open-ended response)
    \item What improvements would you suggest to make BNet better for donors? \\
    (Open-ended response)
\end{enumerate}

\section{Data Analysis}\label{appendix:D}
 To address existing gap of existing BDSs, we ask the following research questions in this work:
\begin{itemize}
    \item \textbf{RQ1:} How can a multi-platform bot be designed to seamlessly integrate with OSNs to accelerate donor response and broaden the donor network?
    \item \textbf{RQ2:} How can a cost-efficient framework be developed to precisely filter blood donation messages from extensive message streams to minimize operational costs?
    \item \textbf{RQ3:} How can a bot serve diverse demographic groups for blood donation and ensure that users perceive its integration as convenient across social media groups?\\
\end{itemize}
To assess our research questions, we formulated three key hypotheses:\\
\textbf{H1} - Auto filtering of blood donation messages and geo-location-based notifications of CBRS will accelerate the speed of donor response.\\
\textbf{H2} - Dual-layered filtering architecture of CBRS will cost-effectively filter and parse blood donation messages from extensive social media streams.\\ 
\textbf{H3} - CBRS, as a multi-platform bot, will serve diverse demographic groups equally and improve convenience across OSNs\\
We first analyzed group messages, blood donation requests per day, and group demographics. For RQ1, we proposed hypothesis H1. To validate H1, we tracked donor response times using timestamps at each stage: message dispatch, bot execution, notification delivery, and response received. For RQ2, we introduced H2, evaluated dual-layered filtering accuracy with precision, recall, and F1-score and assessed the cost-efficiency of this approach. To answer RQ3, we proposed H3, using both quantitative and qualitative analyses of usage logs, pre- and post-study surveys and feedback to assess response quality and satisfaction. Post-study Likert-scale feedback on slash-command prompts, UI design, and user satisfaction, alongside insights from open-ended responses helped highlight improvements and challenges. We demonstrated network growth through multi-platform integration. We also applied Pearson’s correlation and Spearman’s rank correlation and to examine associations between indicators of satisfaction index. A summary of metrics and measures is provided in Table \ref{metric}. Additionally, we reviewed user demographics, compared current blood donation apps, and identified areas for improvement. To minimize message overflow and enhance satisfaction, we explored optimal request frequency, and emphasized security for future implementation.
\subsection{Metrics and Measurements}
To evaluate all hypotheses, we define metrics mentioned in Table \ref{metric}.

% \begin{table*}[t]
% \centering
% \caption{Performance metrics for the evaluation of CBRS.}
% \label{metric}
% \resizebox{\textwidth}{!}{%
% \begin{tabular}{|@{}l|l|l|l@{}|}

% \hline
% \textbf{~Hypothes} & \textbf{Metric} & \textbf{Explanation} & \textbf{Metric System} \\
% \hline

% ~H1 & Timely Response & \begin{tabular}[c]{@{}l@{}}Measurement of the elapsed time between message arrival,\\ request parsing, donor notification and donor response\end{tabular} & Timestamping \\ \hline
% \multirow{}{}{~H2} & Filtering Accuracy & \begin{tabular}[c]{@{}l@{}}Identifying and parsing blood donation requests from \\ a large pool of messages\end{tabular} & \begin{tabular}[c]{@{}l@{}}Precision\\ F1-score\\ Recall\end{tabular} \\
%  & Cost Efficiency & \begin{tabular}[c]{@{}l@{}}The financial implications of filtering and parsing blood \\ donation requests from a large pool of messages\end{tabular} & Pricing Model \\ \hline
% \multirow{}{}{~H3} & Command Usability & Perception of ease of use of the slash command prompt & \multirow{}{}{Likert scale response} \\
%  & Intuitiveness & Perception of user-friendly interface &  \\
%  & Satisfaction Index & Overall user satisfaction in functionality and performance of the bot &  \\ 
%  \hline
 
% \end{tabular}%
% }
% \end{table*}

\begin{table*}[t]
\centering
\caption{Performance metrics for the evaluation of CBRS.}
\label{metric}
\resizebox{\textwidth}{!}{%
\begin{tabular}{|@{}l|l|l|l@{}|}
\hline
\textbf{Hypothesis} & \textbf{Metric} & \textbf{Explanation} & \textbf{Metric System} \\
\hline

H1 & Timely Response & \begin{tabular}[c]{@{}l@{}}Measurement of the elapsed time between message arrival,\\ request parsing, donor notification, and donor response\end{tabular} & Timestamping \\
\hline

\multirow{2}{*}{H2} & Filtering Accuracy & \begin{tabular}[c]{@{}l@{}}Identifying and parsing blood donation requests from\\ a large pool of messages\end{tabular} & \begin{tabular}[c]{@{}l@{}}Precision\\ F1-score\\ Recall\end{tabular} \\
& Cost Efficiency & \begin{tabular}[c]{@{}l@{}}The financial implications of filtering and parsing blood\\ donation requests from a large pool of messages\end{tabular} & Pricing Model \\
\hline

\multirow{3}{*}{H3} & Command Usability & Perception of ease of use of the slash command prompt & \multirow{3}{*}{Likert scale response} \\
& Intuitiveness & Perception of user-friendly interface & \\
& Satisfaction Index & Overall user satisfaction in functionality and performance of the bot & \\
\hline

\end{tabular}%
}
\end{table*}

\paragraph{Timely Response:}
We define Timely Response as the measurement of the time taken from when a message arrives to when it is parsed, a donor is notified and a response is received. We assess it in two ways. First, we use timestamps to track the time from the arrival of the message to the response in each stage. The second method involves measuring satisfaction with the timely response of CBRS in identifying potential donors between October 23 and October 26, 2024, after its integration into groups. Participants respond on a scale from 1 to 5: 5 indicates "Very satisfied," 4 means "Satisfied," 3 means "Neither," 2 means "Dissatisfied," and 1 represents "Very dissatisfied."

\paragraph{Filtering Accuracy:}
We define filtering Accuracy as the ability to identify and extract blood donation requests from a vast array of messages. This assessment incorporates key performance indicators: precision, F1-score, and recall to ensure robust evaluation \cite{45yacouby2020probabilistic}.
%%%avg message count per group-logic given
\paragraph{Cost Efficiency:}
We measure Cost Efficiency as the financial impact of filtering and processing blood donation requests from a large volume of messages. This involves a pricing model that evaluates various message volumes across different groups and analyzes expenses based on existing cost structures. We then compare these costs to the expenses associated solely with parsing blood-related messages.

\paragraph{Command Usability:}
We define Command Usability as the ease with which users can utilize slash commands. To assess this, we posed the question: "How easy do you find using CBRS through slash-command prompts (e.g., /start, /show\_my\_info, etc.)?" Responses are rated on a scale from 1 to 5, where 5 signifies "Extremely easy," 4 indicates "Very easy," 3 denotes "Moderately easy," 2 represents "Slightly easy," and 1 means "Not at all easy."

\paragraph{Intuitiveness:}
We measure Intuitiveness as the perception of the user interface in relation to usability, color scheme, layout, and overall aesthetic appeal. To evaluate this, we ask, "How intuitive do you find the user interface of CBRS?" Respondents rate their experience on a scale from 1 to 5, where 5 represents "Extremely intuitive," 4 indicates "Very intuitive," 3 signifies "Moderately intuitive," 2 denotes "Slightly intuitive," and 1 means "Not at all intuitive."

\paragraph{Satisfaction Index:}
We ask both donors and requesters to evaluate the overall functionality of CBRS with the question, "How would you rate the overall functionality of CBRS?" Responses are rated on a scale from 1 to 5, where 5 represents "Excellent," 4 indicates "Above Average," 3 signifies "Average," 2 reflects "Below Average," and 1 denotes "Very Poor." 

Additionally, we gather user feedback regarding challenges and potential improvements. To assess performance, we inquire, "Do you find CBRS more effective than the blood donation apps or methods you have previously used?" This comparison provides valuable insights into CBRS's effectiveness in enhancing the blood donation experience.

\section{Findings}\label{appendix:E}
In this section, we present our key findings regarding response time, operational cost efficiency, and user convenience in detail. Our analyses reflect the significant reduction in the parsing and retrieval time after deploying CBRS. The dual-layered filtering architecture helps CBRS maintain adequate accuracy while reducing the parsing cost. Our survey results also indicate that CBRS can be helpful for both donors and recipients across diverse demographic groups.

\paragraph{User Evaluation}

% While technical metrics showcase CBRS’s strengths, its real-world value depends on user satisfaction, which we assessed through a detailed study. We surveyed 20 active Telegram communities for feedback.

\begin{figure}[h]
    \centering
    \includegraphics[width=0.4\textwidth]{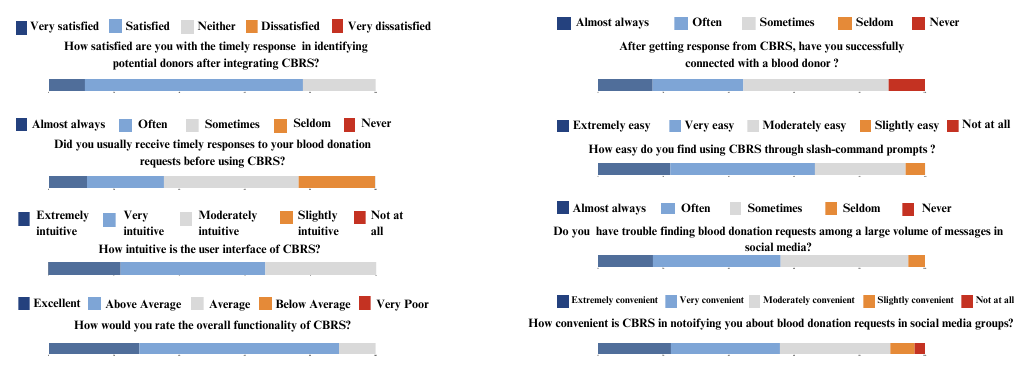}
    \caption{This figure presents the results of a user study conducted through survey questionnaires.}
    \label{survey}
\end{figure}

We conduct a user assessment to evaluate CBRS. User demographics and procedures appear in Appendix \ref{appendix:B}. We recruit members from 20 active Telegram groups and 10 active Discord groups, all based in Bangladesh. User experience-related questions appear in Appendix \ref{appendix:C}. We consider six metrics: Timely Response, Filtering Accuracy, Cost Efficiency, Command Usability, Intuitiveness, and Satisfaction Index, with descriptions provided in Appendix \ref{appendix:D}. We inquire about the overall functionality of CBRS. Notably, 44\% of respondents find command usability to be "very easy," and another 44\% describe the user interface as "very intuitive." Additionally, 61\% rate the overall functionality as "above average," with 28\% considering it "excellent." On social media, 21\% of donors report "always" having trouble finding blood donation requests amid a high volume of messages. Another 32\% experience this "often," and 32\% encounter it "sometimes." After receiving notifications through CBRS, 39\% of donors find the process "very convenient." Figure \ref{survey} shows the results from the survey. We also conduct a Spearman's rank correlation to assess the relationship between user satisfaction and CBRS functionality metrics as shown in Table \ref{tab:spearman}. Command Usability and Intuitiveness show strong positive correlations of 0.52 and 0.54, respectively, with the latter being statistically significant, highlighting the importance of interface design. Timely Notification shows a moderate correlation, while Timely Response shows no significant association.

\begin{table}[h]
\centering
\caption{Result of Spearman Correlation and p-values}
\label{tab:spearman}
\resizebox{\columnwidth}{!}{%
\begin{tabular}{@{}lcc@{}}
\toprule
\textbf{Metric}           & \textbf{Spearman Correlation} & \textbf{p-value} \\
\midrule
Timely Response           & 0.06                          & 0.80              \\
Command Usability         & 0.52                          & 0.02              \\
Intuitiveness             & 0.54                          & 0.01              \\
Timely Notification       & 0.43                          & 0.08              \\
\bottomrule
\end{tabular}%
}
\end{table}

\subsection{H1 Results: Auto-filtering of blood donation messages and geo-location notifications speeds up donor response by reducing parsing and retrieval time }
We track four specific timestamps from message arrival to donor response. First, we log the time the message arrives in the group. Next, we record when the parsed blood donation request is stored in our database; the difference between these two timestamps indicates the time taken to parse the message. Third, we log when a notification is sent to the first matching donor; the time between the second and third timestamps represents the retrieval and matching process. Finally, we capture the first affirmative response from a donor, with the time between the third and fourth timestamps indicating donor response time.
As shown in Table \ref{tab:time}, we observe notable differences at each stage, from message arrival to donor response. Parsing time averaged 4 seconds with a low variability of 0.45 seconds. Retrieval time averaged 5 seconds with a standard deviation of 3 seconds. The most significant finding was Response time, averaging 81 minutes with a high variability of 110 minutes.

\begin{table}[h]
\centering
\caption{Performance of time tracking in each stage of CBRS from arrival to response}
\small
\begin{tabular}{@{}p{0.35\linewidth}p{0.25\linewidth}p{0.3\linewidth}@{}}
\toprule
\textbf{Task} & \makecell[l]{\textbf{Average}\\\textbf{Time}} & \makecell[l]{\textbf{Standard}\\\textbf{Deviation}} \\ \midrule
Parsing Time   & 4s     & 0.45s  \\
Retrieval Time & 5s     & 3s     \\
Response Time  & 81min  & 110min \\
\bottomrule
\end{tabular}
\label{tab:time}
\end{table}

\subsection{H2 Results: The dual-layered filtering architecture of CBRS efficiently filters and parses blood donation messages from large social media streams, delivering high accuracy at a lower cost}
The Layer 1 of the model model was evaluated on the test set, achieving an overall accuracy of 98.7\%. In the classification of messages, non-blood-related messages are denoted as 0 and blood-related messages as 1. As shown in Table \ref{classifier}, for class 0, the model attained a precision of 99\%, with a recall at 99\%, resulting in an F1-score of 0.99. For class 1, precision remained at 99\%, while recall was 98\%, yielding an F1-score of 0.99 as well. Overall, the macro and weighted averages for precision, recall, and F1-score are all 0.99. We also experiment with Logistic Regression using TF-IDF vectorization \cite{24shah2020comparative}. This approach transforms text messages into numerical features by capturing the frequency of unigrams and bigrams. We use L2 regularization to prevent overfitting \cite{25kolluri2020reducing}. We compare this with the DLF model. DLF proves to be the better approach due to its superior handling of bilingual and mixed-language texts. 
\begin{table}[h]
    \centering
    \caption{Classification report of Layer 1 framework}
    \label{classifier}
    \resizebox{\columnwidth}{!}{%
    \begin{tabular}{@{}lcccc@{}}
        \toprule
        \textbf{Class} & \textbf{Precision} & \textbf{Recall} & \textbf{F1-Score} & \textbf{Support} \\
        \midrule
        0 & 0.99 & 0.99 & 0.99 & 249 \\
        1 & 0.99 & 0.98 & 0.99 & 276 \\
        \midrule
        Macro Avg    & 0.99 & 0.99 & 0.99 & 525 \\
        Weighted Avg & 0.99 & 0.99 & 0.99 & 525 \\
        \bottomrule
    \end{tabular}%
    }
\end{table}

We, furthermore, analyzed the cost efficiency of single-layered filtering (using only GPT-4-0-mini) compared to dual-layered filtering (using the CBRS architecture) as shown in Table \ref{cost}. We first recorded the daily message volume from our observed groups per day. Next, we logged the number of blood donation requests identified by CBRS. Using GPT-4-0-mini at a rate of \$0.0003 per message, direct processing costs would be \$0.0045, \$0.0165, and \$0.0285 per day for average message counts of 15, 55, and 95, respectively.  In the initial layer, CBRS filters messages with 98.7\% accuracy, isolating blood donation requests with average counts of 1, 3, and 5 per day. These filtered messages then proceed to the second layer, where GPT-4o-mini performs validation and parsing at a cost of \$0.003, \$0.009, and \$0.015, respectively. Overall, this dual-layered architecture reduces costs by approximately 33.33\% to 47.37\%, depending on message volume.

\begin{table}[h]
\centering
\caption{Cost analysis of dual-layered filtering}
\label{cost}
\resizebox{\columnwidth}{!}{%
\begin{tabular}{@{}lllll@{}}
\toprule
% \textbf{Range} & \textbf{Avverage Messages} & \textbf{Blood Messages} & \textbf{Average Cost} & \textbf{Average Cost of Messages} \\
\textbf{Range} 
& \makecell[l]{\textbf{Average}\\\textbf{Messages}} 
& \makecell[l]{\textbf{Blood}\\\textbf{Messages}} 
& \makecell[l]{\textbf{Average}\\\textbf{Cost}} 
& \makecell[l]{\textbf{Average Cost}\\\textbf{of Messages}} \\
\midrule
0--30     & 15  & 1  & \$0.0045 & \$0.0003 \\
40--70    & 55  & 3  & \$0.0165 & \$0.0009 \\
80--110   & 95  & 5  & \$0.0285 & \$0.0015 \\
\bottomrule
\end{tabular}%
}
\end{table}

\subsection{H3 Results: CBRS will serve diverse demographic groups equally and improve convenience across OSNs through timely notifications, timely responses, command usability, intuitiveness}
Our survey shows diverse demographics concerning gender, age, education, and occupation as shown in Figure \ref{fig:Gender}, \ref{fig:Age}, \ref{fig:Education}, and \ref{fig:Occupation}. Among the participants, 65\% identified as male and 35\% as female. All age groups were represented, with individuals aged 18-33 showing the highest interest in blood donation. Males are more inclined to donate blood than females. Among different professionals, students constituted 30\% of respondents while NGO workers made up 25\% ranking second. Notably, 80\% of the participants had college or undergraduate education. Among these groups, 11\% "almost always" make donation requests, 39\% "seldom" request and 22\% "sometimes" request. However, only 11\% reported receiving timely responses "always". After integrating CBRS, 67\% of users were "satisfied" with the timely responses from CBRS.

% \begin{figure}[h]
%     \centering
%     \includegraphics[width=0.5\textwidth]{survey.pdf}
%     \caption{This figure shows overall survey result of survey.}
%     \label{diagram2}
% \end{figure}

% We inquired about the slash-command prompts, user interface, and overall functionality of CBRS. Notably, 44\% of respondents found command usability to be "very easy" while another 44\% described the user interface as "very intuitive". Additionally, 61\% rated the overall functionality as "above average" with 28\% considering it "excellent". In social media, 21\% of donors report "always" having trouble finding blood donation requests amid a high volume of messages, while 32\% experience this "often," and another 32\% encounter it "sometimes." However, after receiving notifications through CBRS, 39\% of donors find the process "very convenient," and an additional 39\% rate it as "moderately convenient." Figure \ref{UIs} and Figure \ref{figUI} shows full result of different metrics from survey.

\begin{figure}[h]
    \centering
    \includegraphics[width=0.5\textwidth]{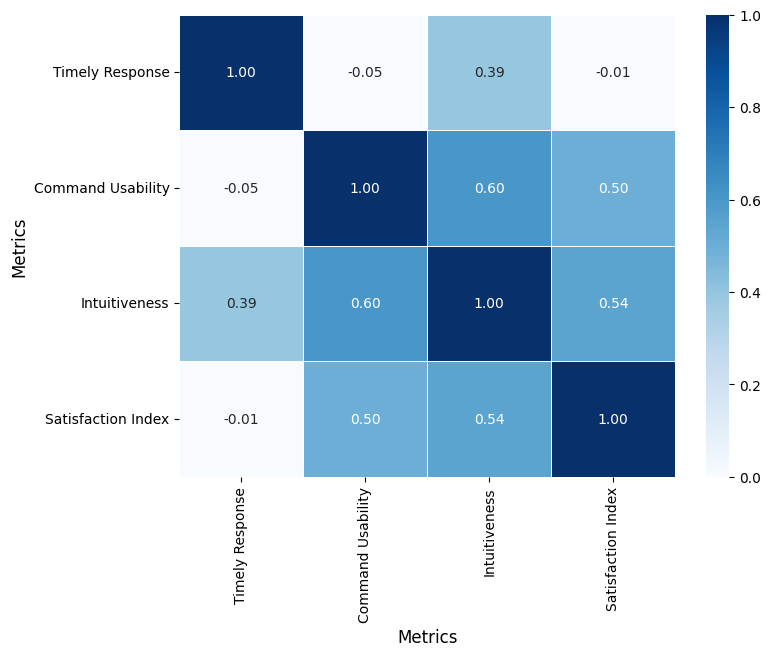}
    \caption{This figure shows Pearson Correlation Heatmap of User Feedback Metrics}
    \label{fig:pearson}
\end{figure}

We first examined the correlations among four metrics—Timely Response, Command Usability, Intuitiveness, and Satisfaction Index—using Pearson’s correlation analysis to explore their interrelationships as shown in Figure \ref{fig:pearson}. Notably, Command Usability and Intuitiveness show a high correlation coefficient of 0.60. Additionally, there is a moderate positive correlation of 0.54 between Intuitiveness and the Satisfaction Index. Command Usability and the Satisfaction Index exhibit a positive correlation of 0.50. However, Timely Response did not significantly correlate with the other metrics, particularly with the Satisfaction Index (-0.01) and Command Usability (-0.05).

However, only 17\% of users reported being "almost always" receiving blood, while 44\% experienced "sometimes", and 11\% never connected even after receiving responses from donors through CBRS. When asked about the challenges they faced while connecting, one donor P33 highlighted, “I encountered communication and transport issues even if I responded to donate”. P54 expressed, “My family did not allow to donate blood to individuals I did not know”. Additionally, P60 remarked, “The location of donation requests was unclear”. We inquired about existing blood donation apps or methods that participants had used previously. Notably, 50\% stated that CBRS is "much better", while 39\% described it as "somewhat better". When asked for suggestions for improvement, P51 remarked, “It would be beneficial to incorporate CBRS into other social media platforms for wider accessibility”. Another participant, P42, suggested, “A dedicated dashboard displaying donation requests would be helpful for users”.

\section{Discussion}\label{appendix:F}
In this study, we designed and developed a multi-platform bot to engage users and efficiently screen large volumes of messages. To keep filtering cost-effective, we implemented a dual-layered filtering architecture. Our evaluation with 114 users showed improved response times and engagement and provided more effective support to existing BDSs. Automated filtering and notifications enabled faster responses, while multi-platform integration created a versatile donor network. Convenient slash commands and an intuitive interface made it easy for participants to use. This section discusses the implications of our findings and provides design recommendations.
\subsection{Creating a fast response system architecture for a multi-platform bot}
There exists a delicate balance between success and failure in urgency management \cite{32joshi2017drasb}. Timely and precise execution during emergencies significantly enhances the likelihood of successful outcomes and mitigates potential risks \cite{32joshi2017drasb}. Auto-notification system is necessary for system administrators in this regard \cite{33aziz2010proactive}. It provides real-time updates on system status and enables prompt responses to issues by maintaining optimal operational efficiency \cite{33aziz2010proactive}. 

Our findings demonstrate that auto-filtering of blood donation messages and geo-location notifications significantly accelerates donor responses. The average parsing time is just 4 seconds with minimal variability. This rapid message filtering enables us to sift through large pools of social media messages and quickly identify relevant donation requests. The retrieval time averages 5 seconds, streamlining the matching process. These data are mentioned in Table \ref{tab:time}. This ensures potential donors are connected to appropriate requests without delay. Auto-notifications use the Haversine distance algorithm to alert nearby donors based on their geographical location. The importance of these features cannot be overstated. This targeted approach minimizes the time and effort needed to locate suitable donations. Response time is a crucial factor we observe. While it averages 81 minutes and shows high variability, it can be significantly influenced by the efficiency of previous steps. Quick parsing and retrieval times help minimize overall delays and facilitates faster connections between donors and recipients.

Before implementing CBRS, we asked users if they received timely responses to their blood donation requests. Many reported they did not get timely replies in Figure \ref{survey}. However, after using CBRS, users expressed satisfaction with the promptness of the responses. When we inquired about difficulties in locating blood donation requests among a large pool of messages, results from our study showed that most struggled to find them from a broad range of messages. In contrast, a significant majority found CBRS effective in notifying them about donation requests within social media groups. Our analysis also revealed that even though donors responded, many were unable to complete the donation. When we asked open-ended questions on this issue, we noted that most donors cited unclear donation location addresses as a primary challenge. We also observed that some donors expressed concerns about donating to unfamiliar recipients. It underscores a potential need for member authentication. We plan to broaden our research to tackle these concerns. We invite researchers from HCI to collaborate on finding alternative solutions for these challenges.

\subsection{Designing cost-optimized dual-layered filtering for free multi-platform Use}
 80\% of people utilize social media for interactions with friends, family, spouses, co-workers, old acquaintances, and new friends \cite{28whiting2013people}. Additionally, 76\% turn to these platforms to pass the time, often during idle moments or when seeking entertainment \cite{28whiting2013people}. Importantly, social media has assumed an increasingly vital role in emergencies \cite{29lindsay2011social}, ranking as the fourth most popular source for accessing emergency information \cite{29lindsay2011social}. Recent advancements in state-of-the-art Large Language Models (LLMs), such as the GPT series, have showcased exceptional reasoning capabilities across various tasks, including message filtering and parsing \cite{30huang2024text}. However, the continuous deployment of LLMs on large message pools can lead to significant operational costs. As illustrated in Figure \ref{fig:Group_member}, group message volume varies with group size, emphasizing the necessity for primary filtering to effectively manage this substantial influx of messages. Ensuring high accuracy during the filtering process is paramount. To strike a balance between cost-effectiveness and accuracy, we implemented a dual-layered structure, resulting in high precision, F1 scores, and recall rates. This approach not only enhances the efficiency of message processing but also minimizes resource expenditure. Our dataset was meticulously curated to provide a balanced representation of positive and negative messages across various languages. It addresses both class imbalance and linguistic diversity over a wide array of topics beyond emergency contexts. The versatility of this dataset is crucial for training models that can generalize effectively in real-world scenarios. Notably, our calculations in Table \ref{cost} indicate that primary filtering can reduce bot operation costs by up to 47\%, as only blood donation requests advance to GPT-4-mini for further processing. This significant cost savings underscores the importance of efficient filtering mechanisms in optimizing the functionality of LLMs in social media applications. Integrating a dual-layered filtering approach can greatly enhance the management of large volumes of messages on social media platforms, particularly in emergency contexts. This framework not only ensures timely and accurate responses but also demonstrates the potential for significant cost reduction, paving the way for more efficient use of advanced language processing technologies.

\subsection{Developing a multi-platform solution for diverse demographics}
Association for the Advancement of Blood \& Biotherapies (AABB) reports that the average blood donor is typically college-educated and aged 30–50 years. Younger adults, particularly those aged 18-25, are increasingly likely to donate blood \cite{35americasbloodcenters2024}. While males have historically been more frequent donors than females \cite{34zambon2020}, this gap is narrowing as more females become regular donors. Additionally, white individuals tend to donate at higher rates compared to Black, Hispanic, and Asian populations \cite{35americasbloodcenters2024}. Blood donors from higher socioeconomic backgrounds are more likely to donate, often due to better access to healthcare facilities and donation centers \cite{34zambon2020}. Table \ref{tab:zgrp} illustrates that the age group most likely to donate blood is also among the most active on social media platforms. This demographic overlap is significant, as males are slightly more active on social media (53.4\%) compared to females (46.6\%). The similarities in demographics between blood donors and social media users indicate that social media can be a powerful tool for identifying potential donors across various demographic groups. Integrating bots on social media platforms can effectively trace and engage potential donors from all demographic categories. 

Our result shows that using a multi-platform approach greatly broadens the donor network by engaging diverse demographics across popular platforms. Each platform offers unique strengths. Telegram is ideal for exchanging messages, sharing media and files, and supporting private or group calls \cite{36yinka2018telegram}. Facebook focuses on connecting communities \cite{38rauch2013advancing}. It is effective for creating and maintaining support groups that foster awareness and keep people updated on ongoing donation needs \cite{38rauch2013advancing}. Discord, initially popular for gaming, allows for real-time text, voice, and video communication in community-centered "servers" \cite{37kruglyk2020discord}. This feature helps reach younger, tech-savvy users \cite{37kruglyk2020discord}. Our survey highlighted that not all blood types are equally available. Rare types like O- and AB- are often harder to find. Limiting the donor search to a single platform would risk missing donors who frequently use other social spaces. By adopting a multi-platform strategy, we increase the probability of reaching donors with diverse blood types and availability. Our survey results also confirmed that a multi-platform approach increases the donor pool. 

When asked for feedback on areas for improvement, participants suggested extending CBRS to other social media channels such as WhatsApp and Facebook. This aligns with our future research plans to integrate more platforms and ensure wider coverage.

% \begin{table}[h!]
% \centering
% \caption{This table shows social media user by different age group \cite{46khoros2024}}
% \begin{tabular}{|c|c|c|}
% \hline
% \textbf{Age Group} & \textbf{Age Range} & \textbf{Social Media Users (millions)} \\ \hline
% Gen Z             & 11-26              & 56.4                                     \\ \hline
% Gen X             & 43-58              & 51.8                                     \\ \hline
% Baby Boomers      & 59-77              & 36.9                                     \\ \hline
% \end{tabular}

% \label{table:age-group-social-media-usage}
% \end{table}

\begin{table}[h]
\centering
\caption{Social media users by different age groups}
\small
\begin{tabular}{@{}p{0.35\linewidth}p{0.25\linewidth}p{0.3\linewidth}@{}}
\toprule
\makecell[l]{\textbf{Age}\\\textbf{Group}} & \makecell[l]{\textbf{Age}\\\textbf{Range}} & \makecell[l]{\textbf{Social Media}\\\textbf{Users}} \\ \midrule
Gen Z        & 11--26 & 56.4M \\
Gen X        & 43--58 & 51.8M \\
Baby Boomers & 59--77 & 36.9M \\
\bottomrule
\end{tabular}
\label{tab:zgrp}
\end{table}

\subsection{Exploring slash command prompt and user Interface design for multi-platform bot}
The user interface of bots are often referred to as the "universal UI" due to their flexibility and ease of use across multiple platforms \cite{31santhanam2022bots}. Integrating command prompt mechanisms into these systems has tremendously enhanced their utility \cite{31santhanam2022bots}. This enhancement facilitates quicker task completion and reduces the need for extensive documentation \cite{31santhanam2022bots}. In our findings, we explored how these design choices influenced overall user satisfaction within CBRS. We asked users about their perception of command usability and the user interface of the bot shown in Figure \ref{survey}. Most reported satisfaction with the performance of CBRS. Our analysis demonstrated that command usability and interface intuitiveness play a pivotal role in fostering a positive user experience. Pearson’s correlation analysis revealed strong relationships between Command Usability, Intuitiveness, and Satisfaction Index which indicates that an intuitive, accessible interface is crucial for user engagement. Spearman’s rank correlation further confirmed these insights by showing a consistently positive relationship between command usability and user satisfaction. As command usability improved, satisfaction levels also increased and intuitive design elements in the interface had a significant statistical impact. This analysis shows that a quality user interface and easy command access are key for a smooth user experience on CBRS.

In our design phase, we selected Telegram and Discord due to their engaging conversational interfaces and shared features of slash commands. We explored these platforms to enhance interaction efficiency. By opting for a structured command interface rather than real-time natural language parsing, we aimed to reduce miscommunication and increase response speed. This design choice made the bot more accessible and user-friendly, ultimately resulting in higher satisfaction levels. To further streamline the experience, we developed a single-page web application that simplifies donor data entry. This application captures essential information such as blood group, last donation date, and GPS location directly from the user’s browser. We observed that this addition, combined with unique URLs linked to users’ chat platform identities, allowed donors to update their details effortlessly without needing to re-identify themselves. These design decisions facilitated seamless information management and had a positive impact on satisfaction, as users could easily update their information from the chat interface. Each feature we implemented, such as simplified data entry and unique URLs, contributed significantly to user satisfaction by enhancing usability and reducing friction. We explored how a multi-platform command prompt and user interface enhance user interactions by ensuring consistent access and intuitive navigation across various platforms. When we solicited open-ended feedback regarding potential improvements for CBRS, users highlighted the need for a dashboard displaying donation requests. This feedback reflects a strong desire for more organized and accessible information. We plan to delve deeper into this feedback to refine and elevate the user experience.

\section{Examples of Misclassified and Misparsed Samples}

Table~\ref{tab:misclassified-examples} presents representative failure cases from the classification pipeline, illustrating false positives and false negatives produced by different classifiers. Tables~\ref{tab:misparsed-1-hospital}, \ref{tab:misparsed-2-time-date}, and \ref{tab:misparsed-3-compensation} present recurring parsing errors including hallucinated fields, date normalization failures, and untranslated output across multiple models and languages.

\begin{table*}[t]
\centering
\small
\caption{Representative misclassified examples across different classifiers.}
\label{tab:misclassified-examples}
\begin{tabularx}{\textwidth}{@{}Xcccl@{}}
\toprule
\textbf{Message} & \textbf{Ground Truth} & \textbf{Predicted} & \textbf{Classifier} \\
\midrule
Its your blood that can save another life. Dear friends, the aim of this Group is to help ourselves through Facebook when we are in need. Anybody from anywhere can post his/her blood urgency here and i wish we ourselves will come up with whatever we have\ldots After all, WE LOVE OUR FRIENDS \& FAMILY WITH EVERYTHING. &
false & true & Word2Vec + SVM \\
\midrule
Apnar blood group ki? Amar blood group A+. Apni ki blood dite ichchhuk? Prochur blood-er request ashteche, location ta janaben. &
false & true & ParaMiniLM + Logistic Regression \\
\midrule
Amar rokter group B+. Chokbazar Medical chara je kono sthane rokto donate korte raji achi. Number: 018XXXXXXX (<NAME>). Karo proyojon hole amar ID ebong number take mention kore din. &
false & true & MiniLM12 + Random Forest \\
\midrule
Keo ki (B-) B negative blood donate korte parba. Monday ratre operation. &
true & false & MiniLM12 + Random Forest \\
\bottomrule
\end{tabularx}
\end{table*}

\begin{table*}[t]
\centering
\small
\caption{Representative misparsed example: hallucinated hospital/location and missing condition field.}
\label{tab:misparsed-1-hospital}
\begin{tabularx}{\textwidth}{@{}p{0.18\textwidth}XX@{}}
\toprule
\textbf{Field} & \multicolumn{2}{l}{\textbf{Content}} \\
\midrule
\textbf{Input message} &
\multicolumn{2}{p{0.75\textwidth}}{``Dhaka-r Shahbage'' ekjon operation-er rogir jonno aj (14-06-21) shondha 7 tar moddhe 2 bag ``O negative'' [O-ve] rokter proyojon. [14th Floor, Ward\#12] Jogajog: BARDEM Hospital, phone: 018XXXXXXX (rogir attio).} \\
\midrule
\textbf{Parsed fields} &
{\ttfamily\footnotesize
\textit{blood\_group:} O-\newline
\textit{bags\_needed:} 2\newline
\textit{patient.name:} ---\newline
\textit{patient.gender:} ---\newline
\textit{patient.age\_group:} ---\newline
\textit{condition:} operation\newline
\textit{location:} Dhaka-r Shahbage\newline
\textit{location\_markers:} Dhaka-r Shahbage\newline
\textit{hospital\_name:} BARDEM Hospital\newline
\textit{probable\_day:} 14/06/2021\newline
\textit{probable\_time:} before 19:00\newline
\textit{contacts[0].name:} ---\newline
\textit{contacts[0].numbers:} 018XXXXXXX\newline
\textit{contacts[0].relation:} rogir attio\newline
\textit{compensation.transportation:} ---\newline
\textit{compensation.allowance:} ---
} &
{\ttfamily\footnotesize
\textit{blood\_group:} O-\newline
\textit{bags\_needed:} 2\newline
\textit{patient.name:} ---\newline
\textit{patient.gender:} ---\newline
\textit{patient.age-group:} ---\newline
\textit{condition:} ---\newline
\textit{location:} Kidney Foundation \& Research Institute, Dhaka\newline
\textit{location\_markers:} Dhaka\newline
\textit{hospital\_name:} Kidney Foundation \& Research Institute\newline
\textit{probable\_day:} 14/06/21\newline
\textit{probable\_time:} before 19:00\newline
\textit{contacts[0].name:} ---\newline
\textit{contacts[0].numbers:} 018XXXXXXX\newline
\textit{contacts[0].relation:} ---\newline
\textit{compensation.transportation:} ---\newline
\textit{compensation.allowance:} ---
} \\
\midrule
\textbf{Model / setting} & \multicolumn{2}{p{0.75\textwidth}}{Meta-Llama-3.1-8B-Instruct, few-shot, Bengali} \\
\textbf{Error type} & \multicolumn{2}{p{0.75\textwidth}}{Hallucinated hospital/location; missing condition; date normalization error} \\
\textbf{Comment} & \multicolumn{2}{p{0.75\textwidth}}{The model correctly extracts blood group, bag count, time, and contact number, but hallucinates an entirely different hospital and location (Kidney Foundation \& Research Institute, Dhaka instead of BARDEM Hospital, Shahbag). It also fails to extract the condition (operation) and omits the donor relation field. The date is unnormalized (14/06/21 vs.\ 14/06/2021).} \\
\bottomrule
\end{tabularx}
\end{table*}

\begin{table*}[t]
\centering
\small
\caption{Representative misparsed example: hallucinated probable time and date normalization error.}
\label{tab:misparsed-2-time-date}
\begin{tabularx}{\textwidth}{@{}p{0.18\textwidth}XX@{}}
\toprule
\textbf{Field} & \multicolumn{2}{l}{\textbf{Content}} \\
\midrule
\textbf{Input message} &
\multicolumn{2}{p{0.75\textwidth}}{\#Delhi 2 Units AB-ve \#Blood \#urgent \#need at AIIMS Hospital \#Jun\_21 Pls Call 981XXXXXXX or 724XXXXXXX @TajinderBagga @nistula @upma23 \#BloodMatters} \\
\midrule
\textbf{Parsed fields} &
{\ttfamily\footnotesize
\textit{blood\_group:} AB-\newline
\textit{bags\_needed:} 2\newline
\textit{patient.name:} ---\newline
\textit{patient.gender:} ---\newline
\textit{patient.age\_group:} ---\newline
\textit{condition:} ---\newline
\textit{location:} AIIMS Hospital\newline
\textit{location\_markers:} Delhi\newline
\textit{hospital\_name:} AIIMS Hospital\newline
\textit{probable\_day:} 21/06\newline
\textit{probable\_time:} ---\newline
\textit{contacts[0].name:} ---\newline
\textit{contacts[0].numbers:} 981XXXXXXX, 724XXXXXXX\newline
\textit{contacts[0].relation:} ---\newline
\textit{compensation.transportation:} ---\newline
\textit{compensation.allowance:} ---
} &
{\ttfamily\footnotesize
\textit{blood\_group:} AB-\newline
\textit{bags\_needed:} 2\newline
\textit{patient.name:} ---\newline
\textit{patient.gender:} ---\newline
\textit{patient.age-group:} ---\newline
\textit{condition:} ---\newline
\textit{location:} AIIMS Hospital\newline
\textit{location\_markers:} AIIMS Hospital\newline
\textit{hospital\_name:} AIIMS Hospital\newline
\textit{probable\_day:} Jun\_21\newline
\textit{probable\_time:} before 24:00\newline
\textit{contacts[0].name:} ---\newline
\textit{contacts[0].numbers:} 981XXXXXXX, 724XXXXXXX\newline
\textit{contacts[0].relation:} ---\newline
\textit{compensation.transportation:} ---\newline
\textit{compensation.allowance:} ---
} \\
\midrule
\textbf{Model / setting} & \multicolumn{2}{p{0.75\textwidth}}{Meta-Llama-3.2-3B-Instruct, few-shot, English} \\
\textbf{Error type} & \multicolumn{2}{p{0.75\textwidth}}{Hallucinated probable time; date normalization error; incorrect location marker} \\
\textbf{Comment} & \multicolumn{2}{p{0.75\textwidth}}{The model correctly extracts blood group, bag count, hospital name, and contact numbers, but hallucinates a probable time (before 24:00) that has no basis in the input. It also fails to normalize the date (Jun\_21 instead of 21/06) and incorrectly uses the hospital name as the location marker instead of the city (Delhi).} \\
\bottomrule
\end{tabularx}
\end{table*}

\begin{table*}[t]
\centering
\small
\caption{Representative misparsed example: incorrect compensation fields, hallucinated time, and untranslated/unnormalized output fields.}
\label{tab:misparsed-3-compensation}
\begin{tabularx}{\textwidth}{@{}p{0.18\textwidth}XX@{}}
\toprule
\textbf{Field} & \multicolumn{2}{l}{\textbf{Content}} \\
\midrule
\textbf{Input message} &
\multicolumn{2}{p{0.75\textwidth}}{Assalamu Alaikum. Esho kori rokto dan, hashbe rogi, bachbe pran. Ajkei dorkar, pliz je dite ichchhuk doya kore ektu jogajog korben. Rogir shomossha: khubi mumurshu obosthay medical-e vorthi ache (boyoshko). Rokter group: O-. Rokter poriman: onirnito (ekadhik lagbe). Rokto daner tarikh: joto druto shombhob, 30-10-2022. Rokto daner sthan: Rajshahi Medical College (alochona kore je kono jagha theke neya jabe). Jogajog: 017XXXXXXX.} \\
\midrule
\textbf{Parsed fields} &
{\ttfamily\footnotesize
\textit{blood\_group:} O-\newline
\textit{bags\_needed:} ---\newline
\textit{patient.name:} ---\newline
\textit{patient.gender:} ---\newline
\textit{patient.age\_group:} adult\newline
\textit{condition:} very critical, admitted in medical\newline
\textit{location:} Rajshahi Medical College\newline
\textit{location\_markers:} Rajshahi\newline
\textit{hospital\_name:} Rajshahi Medical College\newline
\textit{probable\_day:} 30/10/2022\newline
\textit{probable\_time:} ---\newline
\textit{contacts[0].name:} ---\newline
\textit{contacts[0].numbers:} 017XXXXXXX\newline
\textit{contacts[0].relation:} ---\newline
\textit{compensation.transportation:} ---\newline
\textit{compensation.allowance:} ---
} &
{\ttfamily\footnotesize
\textit{blood\_group:} O-\newline
\textit{bags\_needed:} onirnito (ekadhik lagbe)\newline
\textit{patient.name:} ---\newline
\textit{patient.gender:} M\newline
\textit{patient.age\_group:} adult\newline
\textit{condition:} khubi mumurshu obosthay medical-e vorthi ache\newline
\textit{location:} Rajshahi Medical College\newline
\textit{location\_markers:} Rajshahi\newline
\textit{hospital\_name:} Rajshahi Medical College\newline
\textit{probable\_day:} 30/10/2022\newline
\textit{probable\_time:} in as soon as possible\newline
\textit{contacts[0].name:} ---\newline
\textit{contacts[0].numbers:} 017XXXXXXX\newline
\textit{contacts[0].relation:} ---\newline
\textit{compensation.transportation:} N\newline
\textit{compensation.allowance:} N
} \\
\midrule
\textbf{Model / setting} & \multicolumn{2}{p{0.75\textwidth}}{Claude-3-Haiku, zero-shot, Bengali} \\
\textbf{Error type} & \multicolumn{2}{p{0.75\textwidth}}{Incorrect compensation fields; hallucinated probable time; untranslated/unnormalized condition and bags\_needed; hallucinated patient gender} \\
\textbf{Comment} & \multicolumn{2}{p{0.75\textwidth}}{The model correctly extracts blood group, hospital name, location, and date, but makes several errors: it fills compensation fields with ``N'' instead of leaving them blank (compensation was not mentioned in the input); it hallucinates a probable time (``in as soon as possible'') and a patient gender (M) not stated in the input; it copies the raw Bengali text for condition and bags\_needed rather than normalizing them to English canonical forms.} \\
\bottomrule
\end{tabularx}
\end{table*}

\end{document}